\documentclass[11pt]{article}
\usepackage[letterpaper,left=1in,right=1in,top=0.92in,bottom=0.92in]{geometry}
\usepackage[T1]{fontenc}
\usepackage[utf8]{inputenc}
\usepackage{amsmath}
\usepackage{amsthm}
\usepackage{newtxtext}
\usepackage{newtxmath}
\usepackage{bm}
\usepackage{graphicx}
\usepackage[table]{xcolor}
\usepackage{algorithm}
\usepackage{algpseudocode}
\usepackage{booktabs}
\usepackage{multirow}
\usepackage{adjustbox}
\usepackage{float}
\usepackage{placeins}
\usepackage{makecell}
\usepackage{tikz}
\usetikzlibrary{arrows.meta,positioning,backgrounds,calc,shapes.geometric,fit,shapes,chains}
\usepackage[font=small,labelfont=bf,labelsep=quad]{caption}
\usepackage{natbib}
\usepackage[hidelinks]{hyperref}

\newcommand{\real}{\mathbb{R}}

\newcommand{\expect}{\mathbb{E}}
\newcommand{\setx}{\mathcal{X}}
\newcommand{\sety}{\mathcal{Y}}
\newcommand{\xT}{{\top}}
\newcommand{\name}[1]{\texttt{#1}}
\newcommand{\stepa}{\textbf{A}}
\newcommand{\stepb}{\textbf{B}}
\newcommand{\gdgss}{\name{Guided-DGSS}}
\let\Zbar\undefined
\newcommand{\Zbar}{\bar{\mathbf{Z}}}
\newcommand{\gDF}{g^{\mathrm{DF}}}
\newcommand{\dsens}{\bar{\bm{\lambda}}}
\newcommand{\dsenspad}{\bar{\bm{\lambda}}_{\zeta}}
\newcommand{\Usub}{\mathbf{U}}
\newcommand{\Sigmacov}{\Sigma}
\newcommand{\Fcomp}{\mathcal{F}}
\newcommand{\Ftilde}{\widetilde{\mathcal{F}}}
\newcommand{\Lobj}{\mathcal{L}_{\mathrm{OBJ}}}
\newcommand{\Lmse}{\mathcal{L}_{\mathrm{MSE}}}
\newcommand{\plainm}{\name{SPSA-DGSS}}

\newcommand{\analytic}{\name{Analytic-DF}}
\newcommand{\isoes}{\name{Unguided-ES}}
\newcommand{\guidedes}{\name{Guided-DGSS-H}}
\newcommand{\guidedsurr}{\name{Guided-DGSS-DF}}

\newcommand{\floatnotes}[1]{\if\relax\detokenize{#1}\relax\else\par\vspace{4pt}{\footnotesize\raggedright #1\par}\fi}
\newcommand{\TABLE}[3]{\caption{#1}\centering #2\floatnotes{#3}}
\newcommand{\FIGURE}[3]{\centering #1\par\caption{#2}\floatnotes{#3}}
\newcommand{\Halmos}{\hfill$\square$}

\newtheoremstyle{boldplain}{6pt plus 3pt minus 2pt}{6pt plus 3pt minus 2pt}{\itshape}{}{\bfseries}{.}{0.5em}{\thmname{#1}\thmnumber{ #2}\thmnote{ (#3)}}
\newtheoremstyle{boldremark}{6pt plus 3pt minus 2pt}{6pt plus 3pt minus 2pt}{\normalfont}{}{\bfseries}{.}{0.5em}{\thmname{#1}\thmnumber{ #2}\thmnote{ (#3)}}
\theoremstyle{boldplain}
\newtheorem{proposition}{Proposition}
\newtheorem{lemma}{Lemma}
\theoremstyle{boldremark}
\newtheorem{remark}{Remark}
\renewenvironment{proof}[1]{\par\smallskip\noindent\textit{#1}\enspace}{\par\smallskip}

\bibpunct[, ]{(}{)}{,}{a}{}{,}

\title{Boiling Down Uncertainty: Deep Generative Scenario Search with Decision-Guided Evolution Strategies for Two-Stage Stochastic Optimization}
\author{Atefeh Hemmati Golsefidi\thanks{Corresponding author.} \and David Pisinger}
\date{}

\begin{document}

\maketitle
\vspace{-2.2em}
\begin{center}
Department of Technology, Management and Economics\\
Technical University of Denmark, 2800 Kgs.\ Lyngby, Denmark\\
\texttt{ahego@dtu.dk}, \texttt{dapi@dtu.dk}
\end{center}

\begin{abstract}
\noindent One of the most common models for handling uncertainty is two-stage stochastic programming. Choosing the right number of scenarios is a trade-off between the uncertainty that can be represented and computational cost. A recent method, Deep Generative Scenario Search, generates synthetic scenarios and improves them by learning from the decisions they induce. The original approach used a black-box gradient estimate, since decisions are piecewise constant across scenarios. This paper provides the missing analytical foundation by deriving the gradient in closed form from recourse duality and an adjoint solve of the regularized reduced problem. The decision-focused direction, with a heuristic direction as a backup when duals are unavailable, defines the sampling subspace of a guided evolution strategy rather than being used as a direct update. For any guiding subspace, the expected update is a descent direction of the smoothed objective, so an inaccurate guide adds variance while the expected step still descends. For six problem families, both guided estimators outperform the baseline. The decision-focused version reduces the mean gap by three-quarters, and on the two families with the most integer variables the reduced sets beat the exact solver's two-hour incumbent in most runs, using a fifth of the time.

\medskip
\noindent\textbf{Keywords:} stochastic programming; scenario reduction; decision-focused learning; evolution strategies; generative models
\end{abstract}

\section{Introduction}\label{sec:intro}

Two-stage stochastic programming is the standard framework for decisions that must be taken before the uncertainty is resolved, and such decisions are common in operations management. Typical applications include installing generation capacity before demand and prices are known, sizing vehicle fleets before orders are received, locating charging infrastructure before adoption patterns are observed, and positioning relief supplies in advance of disasters. In this framework, an initial decision is made before the uncertainty is resolved, and a recourse action is then taken to address the realized outcome at a cost. The framework is well established, but it carries a heavy computational burden. The extensive form contains one copy of the recourse problem for each scenario that represents the uncertainty. When the recourse stage involves integer variables, as with indivisible units, vehicles, or facilities, each scenario adds a mixed-integer subproblem. There is therefore a trade-off between an accurate representation of the uncertainty and a tractable optimization problem.

One classical response is scenario reduction, in which the full set of scenarios is replaced with a much smaller subset before solving the reduced problem. Most existing literature adopts a distribution-driven perspective, selecting the smaller set to closely approximate the full set according to a probabilistic metric \citep{Dupacova2003,heitsch2003scenario}. However, the effectiveness of a reduced set is ultimately determined by the quality of the decision it induces, rather than its fidelity to the original distribution. Two reduced sets that approximate the distribution equally well may lead to substantially different first-stage decisions, and a set that appears unrepresentative in distribution can still yield a nearly optimal decision. This insight has prompted a problem-driven research direction, where scenarios are selected, clustered, or constructed based on their impact on the optimization problem itself \citep{keutchayan2023problem,hewitt2022decision,bertsimas2023optimization}. A recent development, which forms the basis of this paper, represents a shift from selection to synthesis. \emph{Deep Generative Scenario Search} \citep[\name{DGSS};][]{pisinger2026deep} trains a generative model on empirical scenarios and explores its latent space, a low-dimensional representation learned by the model, to identify a small set of synthetic scenarios. In each iteration of its search loop, the reduced problem is solved for the current synthetic set, the resulting first-stage decision is evaluated on the full empirical set, and the generative model is adjusted to favor synthetic sets whose induced decisions perform better. The cost of a decision on the full empirical set is referred to as the full-set objective. Scenarios are thus evaluated based on the decisions they produce, aligning directly with the problem-driven criterion. Since the generative model is trained to replicate empirical scenarios, the synthetic scenarios preserve the data's statistical characteristics. Unlike selected scenarios, these generated scenarios are novel points that are not restricted to the empirical support and can integrate decision-relevant properties from multiple empirical scenarios. In its initial study, \name{DGSS} outperformed twenty-five scenario-reduction methods in full-set evaluation across three NP-hard problem classes. The present paper retains this search loop but enhances its most important step, the adjustment of the generative model, enabling scenario set reduction to be guided by the structure of the optimization problem rather than by random exploration.

However, adjusting the generative model requires the gradient of the full-set objective with respect to the synthetic scenarios, which is not available. The mapping from synthetic scenarios to the first-stage decision involves solving a mixed-integer program, resulting in a decision that is piecewise constant with respect to the scenarios. The derivative of this decision is therefore zero almost everywhere and undefined at the jumps, so it says nothing about which way the scenarios should move, even though every other component of the loop is smooth. \name{DGSS} therefore treats the entire loop as a black box, a function that returns only the full-set cost of a given synthetic set, and estimates the gradient using simultaneous perturbation stochastic approximation \citep[SPSA;][]{spall1992multivariate}. This approach is robust because it requires no structural knowledge of the problem, but it does not use any problem-specific information. The estimator relies on a single random direction, its variance increases with the dimension of the latent space, and a large part of the computation is spent exploring directions where the objective function barely changes.

This paper investigates how much of the structure of an optimization problem can be exploited by such a black-box search without compromising its robustness. The natural candidate is supplied by decision-focused learning and differentiable optimization \citep{amos2017optnet,agrawal2019differentiable,donti2017task,elmachtoub2022smart,kotary2021survey}, which would differentiate through the reduced problem, or through its linear relaxation, and follow the resulting gradient. This paper shows that this gradient is available in closed form at small computational cost, as the recourse solves at each iteration return the dual multipliers that measure the sensitivity of the cost to the decision, and a single adjoint solve of a regularized relaxation measures the sensitivity of the decision to the scenarios. However, this gradient should not be used directly as the search direction. Under integer recourse, it is computed from a linear relaxation, and the integrality gap of this relaxation biases the direction. A method that follows the gradient deterministically carries this bias into every step, and it can stall at a stationary point of the relaxation that is not optimal for the original problem. The computational results of this study indicate that such deterministic descent performs no better than a randomized baseline. The proposed approach instead uses structural information to determine the sampling locations rather than the search trajectory. The analytic gradient, or a computationally cheaper heuristic direction derived from per-scenario costs, defines the sampling subspace for a \emph{Guided Evolutionary Strategies} estimator \citep{maheswaranathan2019guided}. This estimator perturbs the current solution with Gaussian probes and updates based on the evaluated costs, mirroring the process of standard evolution strategies. The guiding information influences only the distribution from which probes are drawn, with most of the perturbation budget allocated to the low-dimensional subspace defined by the guiding directions, while an isotropic component ensures all directions remain accessible. Since the update is computed from observed costs, inaccuracies in the guide affect only the probe distribution, not the sign of the expected step. This is formalized by demonstrating that the expected update is a descent direction of the Gaussian-smoothed objective for any choice of guiding subspace. Consequently, an accurate guide accelerates the search, while an inaccurate guide only reduces sampling efficiency. Bias in the structural information is converted into variance in the estimate, which allows the method to benefit from structural information even when that structure is biased, as is the case under integer recourse.

\subsection{Contributions}\label{sec:contributions}

The contributions of this paper are threefold. First, we identify the single non-differentiable link of the generative scenario search and derive a decision-focused gradient for it in closed form. The gradient combines two sensitivities that the loop provides at small additional cost. The first is the sensitivity of the full-set cost to the decision, which is read from the recourse duals. The second is the sensitivity of the decision to the synthetic scenarios, which is obtained from one adjoint solve of a regularized reduced problem. We give the conditions under which this gradient is exact and characterize how it degrades under integrality and degeneracy, which is what makes it a guide rather than an update. Second, we show how to use such a gradient inside a randomized search without giving up the robustness of the search. The gradient, or an alternative heuristic direction that needs no duals, defines the sampling subspace of a guided evolution strategy, and we prove that the descent guarantee of guided evolution strategies, stated by \citet{maheswaranathan2019guided} for differentiable objectives, extends to the piecewise-constant objective of the search. An inaccurate guide therefore lowers the sampling efficiency while the expected step still descends, so the bias that integrality introduces into the structural information becomes variance of the estimate. Third, we report a computational study on six families of two-stage programs with linear, integer and degenerate recourse. Each of five reduction methods, namely the two guided estimators, the \plainm\ baseline and two controls, is run $162$ times under matched wall-clock budgets, and its decisions are compared with the solution that the exact solver Gurobi 12.0.0 \citep{gurobi2024} obtains on the full problem. Both guided estimators improve on the \plainm\ baseline of \name{DGSS}, while an unguided control with the same probes is not significantly better than the baseline, so the gain comes from the guiding subspace and not from the extra probes. On most families a single synthetic scenario induces a decision close to the one obtained with all 200 empirical scenarios, and on the two families with the most integer variables, where the exact solver does not close the optimality gap of the extensive form within its time limit, that scenario, found in a fraction of the solver's time, induces in most runs a better decision than the one found by the solver. Which of the two guides is stronger follows the recourse structure as the theory anticipates, and the covariance mix selected for each instance reflects the quality of its guide.

The remainder of the paper is structured as follows. Section~\ref{sec:literature} reviews the related literature. Section~\ref{sec:problem} presents the problem statement and the generative scenario search. Section~\ref{sec:theory} derives the decision-focused gradient, shows when it is exact, and introduces the guided estimator along with its descent guarantee. Section~\ref{sec:computational} details the computational study. Section~\ref{sec:conclusion} summarizes the findings, gives practical guidelines, and outlines directions for future research. Extended computational results are provided in the e-companion.

\subsection{Related Literature}\label{sec:literature}

This work is related to three areas of the literature, namely scenario reduction and generation, decision-focused learning, and evolution strategies. Each area is reviewed below, with emphasis on the capabilities and limitations that motivate the design of the proposed method.

\subsubsection{Scenario Reduction and Generation}\label{sec:lit_reduction}
The classical, distribution-driven approach selects a small scenario set that closely approximates the empirical distribution according to a probability metric. \citet{Dupacova2003} established the theoretical framework for this methodology together with forward-selection and backward-reduction heuristics, which \citet{heitsch2003scenario} developed further. \citet{heitsch2007note} refined these methods for two-stage programs, while \citet{rujeerapaiboon2022scenario} showed the limits of what a reduction that matches the distribution can guarantee. Since distributional proximity alone does not ensure high-quality decisions, a problem-driven approach incorporates the optimization problem into the reduction process. For example, \citet{keutchayan2023problem} cluster scenarios using the problem's cost function as a proximity measure, \citet{hewitt2022decision} group scenarios based on the similarity of the decisions they induce, and \citet{bertsimas2023optimization} formulate the reduction as an optimization problem under a problem-dependent divergence that accounts for induced decisions. \citet{henrion2022problem} go one step further and choose the scenarios so that the expected recourse function of the reduced problem approximates that of the full problem as closely as possible, which bounds the error in the optimal value. Learning-based methods further advance this direction. \citet{wu2022learning} propose CVAE-SIP and CVAE-SIPA, which learn conditional scenario representations for two-stage integer programs with a variational autoencoder and cluster in the learned latent space, \citet{wu2024hgcn2sp} propose HGCN2SP, a reinforcement-learning policy trained with solver feedback to select scenario subsets, and \citet{bengio2020learning} train a model to predict a single representative scenario whose deterministic problem yields a good first-stage decision. All three learn offline from a collection of instances of the same problem that were solved beforehand. They therefore require such a collection and spend the computing time to solve it in advance, and they can be applied only to new instances drawn from the same distribution. \name{DGSS} \citep{pisinger2026deep}, in contrast, requires no training on previously solved instances. It is applied directly to the instance to be solved and differs from the approaches above in two further respects. It is \emph{generative}, synthesizing new scenarios that retain the statistical features of the empirical data while combining properties of several empirical scenarios, and it is \emph{iterative}, refining the synthetic set through feedback from the induced decision rather than selecting a set once. In its original study on three NP-hard problem classes, \name{DGSS} was benchmarked against twenty-five scenario-reduction methods, twenty-two classical and sampling-based selectors, such as fast-forward and simultaneous-backward selection \citep{heitsch2003scenario}, importance sampling, Metropolis-Hastings, accept-reject, $K$-means and Monte Carlo sampling, and the three deep-learning reducers CVAE-SIP, CVAE-SIPA and HGCN2SP, and it achieved the smallest gap to the reference and the most consistent results across instances. The present work adopts the \name{DGSS} search loop as its starting point and uses its SPSA estimator as the baseline for comparison. The difference lies in the step that adjusts the generative model, where the random probing of SPSA is replaced by a guided evolution strategy whose sampling subspace is set by the dual and cost information returned by the optimization problem, through the decision-focused gradient or a heuristic direction, while its architecture and the remaining parts of the search loop are kept unchanged.

\subsubsection{Decision-Focused Learning and Differentiable Optimization}\label{sec:lit_dfl}
In this literature, a learned model produces the data of an optimization problem, typically as a forecast of its uncertain inputs, and is trained based on the quality of the decision produced by the optimizer. This approach requires the derivative of the optimizer's solution with respect to its input data. \citet{amos2017optnet} obtain this derivative for quadratic programs by implicit differentiation of the KKT system, and \citet{agrawal2019differentiable} extend this construction to conic programs. \citet{donti2017task} train probabilistic models end to end through downstream optimization, \citet{elmachtoub2022smart} construct a convex surrogate loss for the predict-then-optimize setting, and \citet{berthet2020learning} obtain gradients via stochastically perturbed optimizers. \citet{kotary2021survey} provide a comprehensive survey of the field. Another line of research addresses combinatorial lower-level problems directly. For example, \citet{wilder2019melding} differentiate through a quadratically regularized linear relaxation, \citet{vlastelica2020differentiation} and \citet{ferber2020mipaal} differentiate through a combinatorial or mixed-integer solver directly, and \citet{mandi2024decision} survey this area. In the present work, the learned model is the generative model that produces the synthetic scenarios, with its output forming the data of the reduced problem. The required sensitivity is obtained through the regularized relaxation (described in Section~\ref{sec:dfgrad}), which aligns with the second line of research. This paper adds the recourse-dual factor of a two-stage program and the use of the resulting direction as a sampling subspace rather than as an update. Section~\ref{sec:theory} demonstrates that, when the direction is used as a sampling subspace, the bias that the relaxation introduces into the direction under integrality influences only the sampling efficiency of the search and not its descent.

\subsubsection{Zeroth-Order Optimization and Evolution Strategies}\label{sec:lit_es}
The proposed method builds on this area. Classical two-point estimators for Gaussian-smoothed objectives have been analyzed by \citet{nesterov2017random}. SPSA \citep{spall1992multivariate,bhatnagar2013service} maintains a cost of two evaluations per iteration, regardless of the number of parameters, by probing a single random direction. Evolution strategies, ranging from the original single-parent and population-based schemes \citep{rechenberg1973,schwefel1993evolution} to natural evolution strategies \citep{wierstra2014natural}, CMA-ES \citep{hansen2001cma}, and the parallel large-scale variant known as OpenAI-ES \citep{salimans2017evolution}, instead rely on multiple probes, and most of them adapt the search distribution. Several methods reduce the variance of these estimators by shaping the search distribution using side information, such as structured or orthogonal directions \citep{choromanski2018structured}, a guiding subspace of surrogate gradients \citep{maheswaranathan2019guided}, or the search history \citep[see the survey of][]{li2020evolution}. This family of methods is appropriate for the current problem for three reasons. First, these methods do not require differentiation of the objective, so a piecewise-constant loss does not pose a formal challenge. Second, they optimize the Gaussian-smoothed objective, which serves as a natural surrogate for a discontinuous function. Third, Guided Evolutionary Strategies \citep{maheswaranathan2019guided} incorporate side information solely through the covariance of the probes, while the update relies on evaluated objective values. The third property is the one that the present setting requires, because the decision-focused gradient constructed in this work is informative but inexact under integrality. An estimator that uses this gradient as a step direction would inherit its bias, whereas an estimator that uses it only to allocate samples still descends in expectation, as established for every subspace in Section~\ref{sec:theory}. The contribution to this area is a new source of guidance, the decision-focused gradient of a stochastic program, along with an analysis of the conditions under which this guidance is exact. Gaussian smoothing is also related to continuation methods for nonconvex objectives \citep{mobahi2015link}. The objective optimized here is piecewise constant because the reduced problem yields an integer decision, so its gradient provides no information. In contrast, the smoothed objective is differentiable, and the expected step of the proposed estimators is a descent direction for this quantity at each iteration. Therefore, the guarantees in Section~\ref{sec:theory} pertain to the smoothed objective, while the computational study evaluates the unsmoothed full-set cost of the resulting decision.

\section{Problem Setting and Generative Scenario Search}\label{sec:problem}

This section establishes the background for the remainder of the paper. Section~\ref{sec:setting} introduces the two-stage stochastic program and the scenario-reduction task. It also defines the full-set objective, the cost of a decision evaluated on all empirical scenarios, which is the quality measure used throughout the paper. Section~\ref{sec:pipeline} reviews the deep generative scenario search proposed by \citet{pisinger2026deep}, which formulates scenario reduction as a continuous search over latent centroids.

\subsection{Two-Stage Stochastic Programs and Scenario Reduction}\label{sec:setting}

A combinatorial two-stage stochastic program has a first-stage decision $x\in\setx\subseteq\real^{n_x}$ and, for each scenario, a recourse decision $y\in\sety\subseteq\real^{n_y}$. The set $\setx$ contains the first-stage constraints, including the integrality of any integer first-stage variables. The first-stage cost vector $c\in\real^{n_x}$ is known, while the second-stage data depend on the realized scenario. Uncertainty is described by a finite set $S$ of empirical scenarios, scenario $s$ being a vector $\mathbf w_s\in\real^{n}$ of random entries with probability $\pi_s\in(0,1]$, $\sum_{s\in S}\pi_s=1$, and the problem reads
\begin{equation}\label{eq:prob}
 \begin{array}{ll}
 \min\limits_{x,\,y_s} & c^\xT x + \sum_{s\in S} \pi_s\, q_s^\xT y_s \\[2pt]
 \mbox{s.t.}  & x \in \setx, \\[2pt]
              & y_s \in \sety_s(x), \qquad s \in S
 \end{array}
\end{equation}
Writing the recourse value of a decision $x$ under scenario $s$, with recourse cost vector $q_s\in\real^{n_y}$, as
\begin{equation}\label{eq:recourse}
Q(x,s) = \min_{y_s}\ \{\, q_s^\xT y_s \ :\ y_s\in\sety_s(x) \,\},
\end{equation}
the full-set objective, the quantity every method in this paper ultimately minimizes, is
\begin{equation}\label{eq:Fdef}
F(x)=c^\xT x+\sum_{s\in S}\pi_s\,Q(x,s)
\end{equation}
We assume that $Q(x,s)$ is finite for every $x\in\setx$ and every $s$, either through relatively complete recourse or because an infeasible recourse problem is assigned a large finite cost, so that $F$ is finite on $\setx$. For large $|S|$, the extensive form of \eqref{eq:prob} becomes computationally intractable, particularly when the recourse is NP-hard and each scenario results in a mixed-integer subproblem. Scenario reduction addresses this by replacing $S$ with a set $S'$ where $|S'|=K\ll|S|$, and solving the reduced problem. The resulting decision is subsequently evaluated on the full set, with the quality of the reduction measured by the full-set objective \eqref{eq:Fdef} for that decision. Thus, a reduced set is considered effective when the decision it produces has a low full-set objective, rather than when it closely resembles $S$.

\subsection{Deep Generative Scenario Search}\label{sec:pipeline}

\name{DGSS} \citep{pisinger2026deep} constructs a reduced set $S'$ of $K$ synthetic scenarios by search rather than by selecting a subset of $S$. Without loss of generality, it is assumed throughout that these scenarios have equal weights $1/K$. In a pretraining phase (Step~\stepa), an autoencoder with encoder $E_{\theta_e}:\real^{n}\to\real^{m}$ and decoder $D_{\theta_d}:\real^{m}\to\real^{n}$, with latent dimension $m\ll n$, where $n$ is the dimension of a scenario, is trained to reconstruct the empirical scenarios by minimizing a mean-squared loss $\Lmse$. In the search phase (Step~\stepb), each iteration encodes the empirical scenarios, applies $K$-means in latent space, and forms differentiable centroids $\Zbar=[\bar{\mathbf z}_1,\dots,\bar{\mathbf z}_K]\in\real^{m\times K}$ by holding the cluster assignment fixed, so that $\Zbar$ is a smooth function of the encoder parameters. Decoding the centroids yields the synthetic scenario set
\begin{equation}\label{eq:decode}
S'(\Zbar)=\big\{\,\hat{\mathbf w}'_k=D_{\theta_d}(\bar{\mathbf z}_k)\,\big\}_{k=1}^{K},
\end{equation}
solving the reduced problem on it yields the first-stage decision
\begin{equation}\label{eq:reduced}
\hat x(S')=\arg\min_{x\in\setx}\; c^\xT x+\frac{1}{K}\sum_{k=1}^{K} Q(x,\hat{\mathbf w}'_k),
\end{equation}
and evaluating that decision on the full empirical set yields the objective-guided loss
\begin{equation}\label{eq:obj_eval}
\Lobj = F\big(\hat x(S'(\Zbar))\big).
\end{equation}
Training minimizes $\mathcal L=\Lmse+\Lobj$. The reconstruction component backpropagates through the autoencoder in the standard way, and the remaining task is to compute the gradient of \eqref{eq:obj_eval} with respect to the centroids,
\begin{equation}\label{eq:gz}
g_{\Zbar} := \nabla_{\Zbar}\,F\big(\hat x(S'(\Zbar))\big).
\end{equation}
For brevity, $\Fcomp(\Zbar):=F(\hat x(S'(\Zbar)))=\Lobj$ denotes the full-set cost as a function of the centroids, referred to as the objective $\Fcomp$ throughout this paper. Figure~\ref{fig:pipeline} illustrates the loop, presenting the forward pass from the empirical scenarios to the full-set cost and, along the same path in the reverse direction, the decision-focused gradient constructed in Section~\ref{sec:theory}.

\begin{figure}[t]
\FIGURE
{\begin{adjustbox}{max width=\textwidth}
\begin{tikzpicture}[>=Latex, font=\small,
  box/.style={draw, rounded corners, align=center, inner sep=4pt,
              minimum height=12mm, minimum width=17mm},
  lat/.style={draw, rounded corners, align=center, inner sep=3pt, fill=gray!12,
              minimum height=12mm, minimum width=12mm},
  sub/.style={below=0.5pt, font=\scriptsize, text=gray!55!black}]
  \definecolor{dfB}{RGB}{0,100,0}
  \definecolor{spsaP}{RGB}{13,71,161}
  \node[box] (emp) {Empirical\\scenarios $S$};
  \coordinate (encL) at ($(emp.east)+(0.7,0)$);
  \fill[blue!8,draw=blue!45!black] (encL)+(0,0.9) -- ($(encL)+(1.5,0.32)$)
        -- ($(encL)+(1.5,-0.32)$) -- ($(encL)+(0,-0.9)$) -- cycle;
  \node[font=\scriptsize] at ($(encL)+(0.72,0)$) {$E_{\theta_e}$};
  \node[font=\scriptsize, blue!45!black, align=center] at ($(encL)+(0.72,1.15)$) {encoder};
  \node[lat] (z) at ($(encL)+(2.35,0)$) {$\Zbar$};
  \node[font=\scriptsize, align=center] at ($(z.north)+(0,0.30)$) {$K$-means\\centroids};
  \coordinate (decL) at ($(z.east)+(0.15,0)$);
  \fill[purple!8,draw=purple!50!black] (decL)+(0,0.32) -- ($(decL)+(1.5,0.9)$)
        -- ($(decL)+(1.5,-0.9)$) -- ($(decL)+(0,-0.32)$) -- cycle;
  \node[font=\scriptsize] at ($(decL)+(0.78,0)$) {$D_{\theta_d}$};
  \node[font=\scriptsize, purple!50!black, align=center] at ($(decL)+(0.78,1.15)$) {decoder};
  \node[box, right=3.4cm of z] (syn) {Synthetic\\scenarios $S'$};
  \node[box, right=2.9cm of syn] (dec) {Decision\\$\hat x(S')$};
  \node[box, right=2.9cm of dec] (cost) {Full-set\\cost $F$};
  \draw[->,thick] (emp.east) -- (encL);
  \draw[->,thick] ($(decL)+(1.5,0)$) -- (syn.west);
  \draw[->,thick] (syn) -- (dec) node[sub,midway,align=center]{solve the\\ reduced problem};
  \draw[->,thick] (dec) -- (cost) node[sub,midway,align=center]{evaluate on\\ the full set $S$};
  \coordinate (tcost) at ($(cost.north)+(0.10,1.70)$);
  \coordinate (tz)    at ($(z.north)+(0.28,1.70)$);
  \draw[dotted,very thick,spsaP] ($(cost.north)+(0.10,0)$) -- (tcost);
  \draw[dotted,very thick,spsaP] (tcost) -- (tz)
        node[midway,above=2pt,font=\footnotesize,align=center,spsaP]
        {\textbf{\name{SPSA-DGSS} uses no derivatives.} The loop is treated as a black box.\\
         Perturb $\Zbar\pm\sigma\Delta$ along one random direction, rerun it, and difference the two costs $F$};
  \draw[->,dotted,very thick,spsaP] (tz) -- ($(z.north)+(0.28,0.62)$);
  \coordinate (bcost) at ($(cost.south)+(0,-1.62)$);
  \coordinate (bdec)  at ($(dec.south)+(0,-1.62)$);
  \coordinate (bsyn)  at ($(syn.south)+(0,-1.62)$);
  \coordinate (bz)    at ($(z.south)+(0,-1.62)$);
  \draw[dashed,very thick,dfB] (cost.south) -- (bcost);
  \draw[->,dashed,very thick,dfB] (bcost) -- (bdec)
        node[midway,above=4pt,font=\footnotesize,dfB,align=center]
        {$\dsens=\nabla_x F$\\[-2pt]{\scriptsize recourse duals $\bm{\lambda}_s$}};
  \draw[->,dashed,very thick,dfB] (bdec) -- (bsyn)
        node[midway,above=4pt,font=\footnotesize,dfB,align=center]
        {$\big(\partial\hat x/\partial S'\big)^{\!\xT}$\\[-2pt]{\scriptsize one KKT adjoint solve}};
  \draw[->,dashed,very thick,dfB] (bsyn) -- (bz)
        node[midway,above=4pt,font=\footnotesize,dfB,align=center]
        {$J_D^\xT$\\[-2pt]{\scriptsize decoder Jacobian}};
  \draw[->,dashed,very thick,dfB] (bz) -- (z.south);
  \node[font=\footnotesize,dfB,align=center] at ($(bdec)!0.5!(bsyn)+(0,-0.72)$)
        {\textbf{This paper differentiates the same loop analytically.} The factors compose into the\\
         decision-focused gradient $\gDF_{\Zbar}$, which guides the sampling of the evolution strategy};
\end{tikzpicture}
\end{adjustbox}}
{The generative scenario-search loop and the two approaches for obtaining its gradient. The black middle row depicts the \name{DGSS} loop of \citet{pisinger2026deep}, which is common to both methods. In this loop, an autoencoder $(E_{\theta_e},D_{\theta_d})$ maps empirical scenarios $S$ to latent centroids $\Zbar$ and decodes them into synthetic scenarios $S'$ with $|S'|\ll|S|$. The induced decision $\hat x(S')$ is then evaluated on the full set $S$ to compute the cost $F$. The \plainm\ estimator from previous work (blue) is derivative-free, rerunning the loop at $\Zbar\pm\sigma\Delta$ along a random direction $\Delta$ and differencing the resulting costs. In contrast, this paper (green) analytically differentiates the same loop, assembling the decision sensitivity $\dsens=\nabla_x F$ from the recourse duals $\bm{\lambda}_s$, the reduced-solution sensitivity $(\partial\hat x/\partial S')^{\xT}$ from a single sparse KKT adjoint solve, and the decoder Jacobian $J_D^{\xT}$ (Section~\ref{sec:dfgrad}). The composition of these components, the decision-focused gradient $\gDF_{\Zbar}$, is used not as an update direction but as the guiding subspace for the evolution strategy (Section~\ref{sec:guided}).\label{fig:pipeline}}
{}
\end{figure}

\begin{figure}[t]
\FIGURE
{\includegraphics[width=0.75\textwidth]{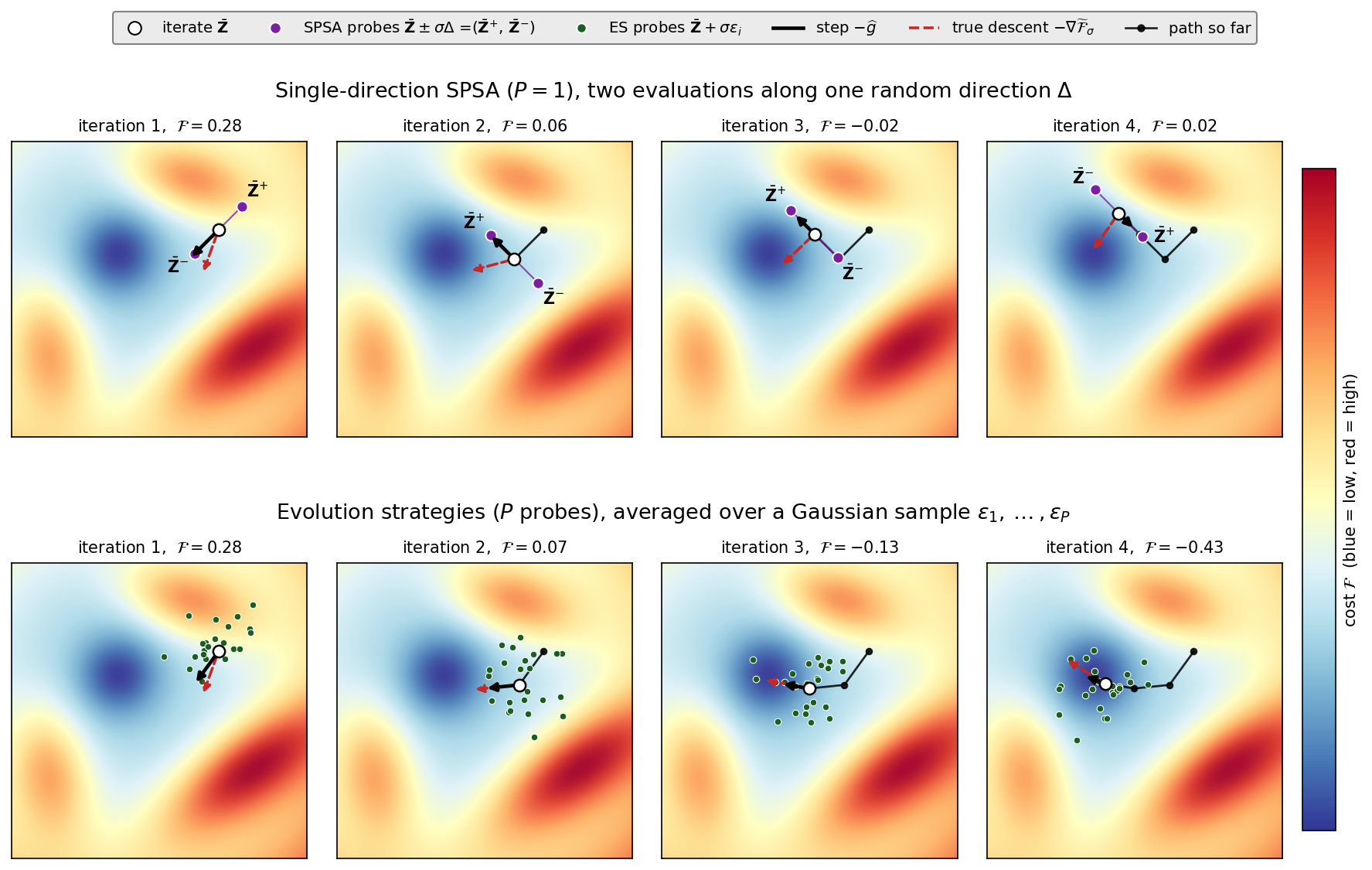}}
{Single-direction SPSA and a multi-probe evolution strategy on a cost surface, four iterations each from the same starting point. The surface represents $\Fcomp$ over two latent coordinates, blue indicates low cost and red high cost, and the red dashed arrow shows the true descent direction. SPSA (top row) evaluates two points along a single random direction, so its estimate reflects the full randomness of that direction, and its fourth step ascends. The evolution strategy (bottom row) averages $P=24$ probes, reducing the variance by a factor of $P$, and reaches the low-cost basin. Both methods use steps of the same decaying length. The surface is drawn smooth for clarity. In the setting of this paper the surface is piecewise constant, and the red dashed arrow represents the gradient of the smoothed objective.\label{fig:estimators}}
{}
\end{figure}

Every link in the chain $\theta\to\Zbar\to S'\to\hat x\to F$, with $\theta=(\theta_e,\theta_d)$, is differentiable except one. The $\arg\min$ of the reduced problem \eqref{eq:reduced} maps scenarios to a first-stage decision. Where this decision is integer, $\hat x$ is piecewise constant in $\Zbar$, and therefore $\Lobj$ is also piecewise constant. Where the first stage has continuous components, $\hat x$ is piecewise smooth with jumps, and the same methods apply. This constitutes the sole non-differentiable link in the loop and is the point at which the methods presented in this paper diverge. \name{DGSS} estimates \eqref{eq:gz} using single-direction SPSA \citep{spall1992multivariate,bhatnagar2013service}. Two evaluations of the complete loop at $\Zbar\pm\sigma\Delta$, with radius $\sigma>0$, along a single random Rademacher direction $\Delta\in\{-1,1\}^{m\times K}$ \citep{Hitczenko1994} yield a finite-difference estimate. This estimate requires nothing from the problem and, when $\Fcomp$ is smooth, matches its gradient in expectation up to a term of order $\sigma^{2}$. However, for the piecewise-constant objective, it represents a finite difference across any discontinuities between the two probes. Employing a single random direction in $d=mK$ latent dimensions provides limited information, resulting in a noisy estimate precisely where evaluations are most costly. Figure~\ref{fig:estimators} contrasts this single-direction probing with the multi-probe sampling characteristic of evolution strategies, the family of estimators on which the methods of this paper are built. The following section introduces this family, shows how problem structure can guide where its probes are sampled, and proves that this guidance is safe in the sense that the expected update remains a descent direction of the Gaussian-smoothed objective for any guiding subspace.

\section{Decision-Guided Search: Structure and Guarantees}\label{sec:theory}

The quality of the synthetic scenarios depends on the effectiveness of the search process, which in turn relies on the accuracy of the gradient estimate used to adjust the generative model. The estimator described in Section~\ref{sec:pipeline} computes this estimate by treating the loop as a black box, resulting in the noise associated with a single random direction. However, the loop is not truly a black box. Each iteration solves the reduced problem on the synthetic scenarios and each recourse subproblem on the full set. These solves yield dual prices when the programs are linear, and the dual prices of their linear relaxations when they are mixed-integer, at little additional cost. Classical sensitivity analysis applies to these underlying programs. Consequently, each iteration generates the information required for an informed search direction, but \name{DGSS} does not utilize this information.

The primary challenge in using this information is its limited reliability. The objective $\Fcomp$ is piecewise constant, so an informative direction can only be obtained from a relaxation of the reduced problem. Under integer recourse, this direction is biased due to the relaxation, and under degenerate duals, it is not uniquely defined. A deterministic update that follows such a direction fully inherits this bias and cannot escape the points to which the bias leads. The central design question is therefore how a search can use structural information without relying on its complete accuracy. The approach developed here assigns distinct roles to the two sources of information. The structural direction determines where the samples are placed, while the evaluated costs of those samples decide how the search moves.

Evolution strategies, an established class of zeroth-order optimization methods \citep{rechenberg1973,schwefel1993evolution}, enable this separation of roles. An evolution strategy generates $P$ random perturbations around the current centroids, evaluates the full-set cost at each perturbed point, and combines these costs into an update. This approach relies solely on function values and is directly applicable to the piecewise-constant objective $\Fcomp$. The single-direction estimator of \name{DGSS} is, up to the perturbation distribution, a special case with $P=1$. Averaging $P$ probes reduces the variance of the estimate by a factor of $P$ (Figure~\ref{fig:estimators}). A key advantage of this family is that probes do not need to be drawn equally in all directions. In a guided evolution strategy \citep{maheswaranathan2019guided}, most of the sampling budget is allocated to a low-dimensional subspace defined by supplied guiding directions, while a fixed portion remains distributed across all directions to ensure none are neglected. This subspace allows structural information to influence probe placement without being fully trusted, as it affects only the distribution of probes and not the update mechanism. An accurate guide focuses probes where the objective varies most, whereas an inaccurate guide merely misplaces them. In both cases, the expected update is a descent direction of the Gaussian-smoothed objective wherever its gradient is nonzero. The closed-form gradient derived in the next subsection, or its computationally cheaper heuristic alternative, provides the guiding directions.

The remainder of this section develops the two key components of this design. Section~\ref{sec:dfgrad} derives the structural direction, presenting a closed-form decision-focused estimate of the gradient \eqref{eq:gz}, and characterizes the conditions under which its two factors are exact. Section~\ref{sec:guided} incorporates this estimate, along with a computationally cheaper heuristic alternative, as the sampling subspace of the guided evolution strategy and proves the descent guarantee for any choice of subspace.

\subsection{A Decision-Focused Gradient in Closed Form}\label{sec:dfgrad}

Each component of the structural direction is already produced by the loop of Section~\ref{sec:pipeline}, specifically the recourse duals from the full-set evaluation, the solution and multipliers of the reduced problem, and the decoder. Applying the chain rule to \eqref{eq:gz}, from the full-set objective back through the solution of the reduced problem on the synthetic scenarios to the centroids, factors the desired gradient into three parts,
\begin{equation}\label{eq:gdf}
\gDF_{\Zbar}
\;=\;
J_D^\xT
\left(\frac{\partial \hat x}{\partial S'}\right)^{\!\xT}
\dsens
\end{equation}
The \emph{decision sensitivity} $\dsens=\nabla_x F(\hat x)\in\real^{n_x}$ measures how the full-set cost responds to the first-stage decision. It is expressed as a dual aggregate, as \eqref{eq:dualgrad} below constructs it from the recourse duals. The \emph{reduced-solution sensitivity} $\partial\hat x/\partial S'\in\real^{n_x\times nK}$ quantifies how the reduced decision changes in response to the synthetic scenarios. The \emph{decoder Jacobian} $J_D=\partial S'/\partial\Zbar\in\real^{nK\times mK}$ maps centroid directions to scenario space, and its transpose pulls the scenario-space gradient back to the centroids. The product with this transpose is obtained by automatic differentiation through the decoder and requires no additional solve. Section~\ref{sec:dualsens} derives the decision sensitivity, Section~\ref{sec:redsens} derives the reduced-solution sensitivity, and Section~\ref{sec:exactness} states when the two sensitivities are exact.

\subsubsection{Decision Sensitivity from Recourse Duality}\label{sec:dualsens}
Although the recourse is re-optimized for each scenario, the envelope theorem indicates that the effect of the re-optimized solution vanishes at the optimum, leaving only the explicit dependence of the constraints on $x$. With the recourse feasible set written as $\sety_s(x)=\{\,y_s\in\real^{n_y} : W_s\,y_s \ge h_s - T_s\,x\,\}$, where the recourse has $n_y$ variables and $r_y$ constraints, so that $W_s\in\real^{r_y\times n_y}$, $T_s\in\real^{r_y\times n_x}$, and $h_s\in\real^{r_y}$, and with $\bm{\lambda}_s\in\real^{r_y}_{+}$ the optimal dual multipliers associated with these constraints, strong duality gives $Q(x,s)=\bm{\lambda}_s^\xT(h_s-T_s\,x)$ and the left part of \eqref{eq:dualgrad}. Substitution in \eqref{eq:Fdef} gives the right part of \eqref{eq:dualgrad}.
\begin{equation}\label{eq:dualgrad}
\makebox[0pt]{$\displaystyle \frac{\partial Q(x,s)}{\partial x}=\frac{\partial}{\partial x}\Big[\bm{\lambda}_s^\xT(h_s-T_s\,x)\Big]=-T_s^\xT\bm{\lambda}_s,\qquad\dsens = c-\sum_{s\in S}\pi_s\, T_s^\xT\bm{\lambda}_s$}
\end{equation}
When the recourse is formulated as a linear program, the dual variables are obtained from the full-set evaluation \eqref{eq:obj_eval} without additional computational cost, and $\dsens$ then follows from the weighted sum over the scenarios in \eqref{eq:dualgrad}. In cases where the recourse is integer, the dual variables correspond to those of the linear relaxation of each recourse problem, which requires one additional linear solve per scenario. This cost is minor compared to the integer solve.

\subsubsection{Reduced-Solution Sensitivity by Implicit Differentiation}\label{sec:redsens}
The integer decision $\hat x$ is piecewise constant in $S'$, so differentiation is performed through the linear relaxation of the reduced problem, which is made strongly convex by the addition of a proximal term with a small weight $\mu>0$. Let $\zeta=(x,y)\in\real^{n_\zeta}$, with $n_\zeta=n_x+Kn_y$, denote the stacked first- and second-stage variables of the reduced problem. The optimal solution of the regularized relaxation is
\begin{equation}\label{eq:relax}
\zeta^\star(S')=\arg\min_{\zeta}\ \big\{\,\tfrac{\mu}{2}\|\zeta\|_2^2+c_\zeta^\xT \zeta\ :\ A(S')\,\zeta\le b(S')\,\big\}
\end{equation}
Here, $\|\cdot\|_2$ is the Euclidean norm, and $c_\zeta\in\real^{n_\zeta}$ represents the concatenation of the first-stage cost $c$ and the weighted recourse costs for the $K$ synthetic scenarios. The matrix $A(S')\in\real^{r_\zeta\times n_\zeta}$ denotes the constraint matrix, and $b(S')\in\real^{r_\zeta}$ is the right-hand-side vector of the reduced problem with $r_\zeta$ constraints, and both depend on the values of the synthetic scenarios. The rows of $A(S')$ include the bounds on the variables, among them the bounds that replace the integrality constraints. An equality constraint of the reduced problem is included as a row that is always active. The Lagrange multiplier of an inequality row is nonnegative, and the multiplier of an equality row is unrestricted in sign. Only the first-stage part of $\zeta^\star$ is needed, since the full-set objective depends on the reduced problem solely through the first-stage decision. In what follows, the reduced-solution sensitivity in \eqref{eq:gdf} therefore refers to that of $\hat x_\mu$, the $x$-block of $\zeta^\star$, which replaces the integer decision $\hat x$ because the latter is piecewise constant with respect to $S'$.

The Jacobian $\partial\hat x/\partial S'$ contains one row for each component of the decision variable and one column for each scenario entry. Constructing this Jacobian requires one linear solve of the size of \eqref{eq:kkt} for each scenario entry. In the gradient expression \eqref{eq:gdf}, the Jacobian appears only within the product $(\partial\hat x/\partial S')^{\xT}\dsens$. With $\dsens$ held fixed, this product is the gradient of a scalar, $(\partial\hat x/\partial S')^{\xT}\dsens=\nabla_{S'}\big(\dsens^\xT\hat x_\mu\big)$. The full Jacobian is therefore not required, and the gradient of this scalar with respect to all scenario entries is obtained from one additional linear solve, as the next paragraph shows.

Let $M\in\real^{r_M\times n_\zeta}$ denote the submatrix of the $r_M$ rows that are active (or tight) at $\zeta^\star$, and let $\nu\in\real^{r_M}$ represent the Lagrange multipliers corresponding to these rows, which the solver provides alongside $\zeta^\star$ without additional computational cost. Let $b_M\in\real^{r_M}$ denote the associated right-hand sides. At the optimum, complementary slackness ensures that the multipliers for all inactive rows are zero. Consequently, the KKT conditions of \eqref{eq:relax} simplify to $\mu\zeta+c_\zeta+M^\xT\nu=0$ and $M\zeta=b_M$. Differentiating these two conditions \citep{fiacco1983introduction,amos2017optnet} yields the single sparse adjoint system
\begin{equation}\label{eq:kkt}
\begin{bmatrix}\mu I & M^\xT\\[2pt] M & 0\end{bmatrix}
\begin{bmatrix}u\\[2pt] p\end{bmatrix}
=
\begin{bmatrix}\dsenspad\\[2pt] 0\end{bmatrix},
\end{equation}
where $\dsenspad\in\real^{n_\zeta}$ equals $\dsens$ from \eqref{eq:dualgrad} on the $x$-coordinates and zero elsewhere. The variables $u\in\real^{n_\zeta}$ and $p\in\real^{r_M}$ are auxiliary adjoint variables, with $u$ containing one entry per component of $\zeta$ and $p$ one entry per active row. Solving \eqref{eq:kkt} once for $(u,p)$ yields the sensitivities of $\dsens^\xT\hat x_\mu$ with respect to every scenario entry in closed form,
\begin{equation}\label{eq:adjoint}
\frac{\partial(\dsens^\xT\hat x_\mu)}{\partial b_i}=p_i,
\qquad
\frac{\partial(\dsens^\xT\hat x_\mu)}{\partial A_{ij}}=-\big(\nu_i\,u_j+\zeta_j\,p_i\big)
\end{equation}
Here, $\nu_i$ and $\zeta_j$ represent the respective components of $\nu$ and the optimal solution $\zeta^\star$. Each iteration therefore requires one dual pass, which is the computation of the recourse duals of all scenarios, one solve of the regularized relaxation \eqref{eq:relax}, and one sparse linear solve to compute the complete decision-focused gradient \eqref{eq:gdf}. Let $g_{S'}=\nabla_{S'}\big(\dsens^\xT\hat x_\mu\big)\in\real^{nK}$ denote the scenario-space gradient, so that $\gDF_{\Zbar}=J_D^{\xT}g_{S'}$. The component corresponding to the scenario entry $S'_t$ is determined by \eqref{eq:adjoint} and depends on the location of that entry,
\begin{equation}\label{eq:vcases}
[g_{S'}]_t=
\begin{cases}
p_i & \text{$S'_t$ in $b_i$,}\\[2pt]
-\big(\nu_i u_j+\zeta_j p_i\big) & \text{$S'_t$ in $A_{ij}$,}
\end{cases}
\end{equation}
where $i$ denotes an active row and the entry enters with coefficient one. An entry that enters with another coefficient multiplies the term by that coefficient, the derivative with respect to an entry in an inactive row is zero, and if an entry appears in multiple positions, the corresponding terms are summed.

\subsubsection{Exactness and Bias of the Gradient}\label{sec:exactness}
The following proposition specifies the conditions under which each of the two sensitivities is exact.

\begin{proposition}[Exactness of the decision and reduced-solution sensitivities]\label{prop:exact}
Consider the gradient \eqref{eq:gdf} constructed from \eqref{eq:dualgrad} and \eqref{eq:adjoint}.
\begin{enumerate}
\item[(a)] If the recourse \eqref{eq:recourse} is formulated as a linear program with a unique optimal dual solution $\bm{\lambda}_s$ at $\hat x$ for each $s\in S$, then $\dsens=\nabla_x F(\hat x)$.
\item[(b)] If the gradients of the active constraints at $\zeta^\star$ are linearly independent and each active inequality constraint possesses a strictly positive multiplier, then $\zeta^\star(\cdot)$ is continuously differentiable in a neighborhood of $S'$, and \eqref{eq:adjoint} is equal to $\nabla_{S'}\big(\dsens^\xT \hat x_\mu(S')\big)$, where $\hat x_\mu$ denotes the $x$-block of $\zeta^\star$.
\end{enumerate}
\end{proposition}

\begin{proof}{Proof of Proposition~\ref{prop:exact}.}
\textbf{Part (a).}
The recourse \eqref{eq:recourse} is a linear program in which $x$ appears exclusively on the right-hand side of its constraints, $h_s-T_s x$. Dual feasibility depends solely on the recourse matrix $W_s$ and the cost vector $q_s$, implying that the dual feasible set $\Lambda_s$ is independent of $x$, and $x$ is present only in the dual objective. By strong duality, it follows that
\begin{equation*}
Q(x,s)=\max_{\bm{\lambda}\in\Lambda_s}\ \bm{\lambda}^\xT(h_s-T_s x)+\psi_s(\bm{\lambda}),
\end{equation*}
where $\psi_s$ represents the dual contribution from any recourse constraints and variable bounds that do not involve $x$. For a fixed $\bm{\lambda}$, the expression inside the maximum becomes $\bm{\lambda}^\xT h_s-\bm{\lambda}^\xT T_s x+\psi_s(\bm{\lambda})$, which is an affine function of $x$ with gradient $-T_s^\xT\bm{\lambda}$ with respect to $x$. Since $Q(\cdot,s)$ is the maximum of finitely many such affine functions, each corresponding to a vertex of $\Lambda_s$, it is convex and piecewise linear in $x$. The function is differentiable at $x=\hat x$ if the maximizer $\bm{\lambda}_s$ is unique at that point, in which case $\partial Q(\hat x,s)/\partial x=-T_s^\xT\bm{\lambda}_s$. Summing over $s$ with weights $\pi_s$ and including the first-stage cost yields $\nabla_x F(\hat x)=c-\sum_{s\in S}\pi_s T_s^\xT\bm{\lambda}_s=\dsens$.

\textbf{Part (b).} The proof is structured in three steps.
Step 1 establishes the differentiability of the solution map.
For $\mu>0$, the objective of \eqref{eq:relax} is strongly convex, ensuring that $\zeta^\star$ is unique. According to the sensitivity theorem of \citet{fiacco1983introduction}, under the conditions of linear independence and strict complementarity, the active set remains the same for all scenario values in a neighborhood of $S'$, and $(\zeta^\star,\nu)$ is continuously differentiable with respect to the data $A(S')$ and $b(S')$ in this neighborhood. On the fixed active set, the solution is characterized by the two KKT equalities $\mu\zeta+c_\zeta+M^\xT\nu=0$ and $M\zeta=b_M$. The Jacobian of these equalities with respect to $(\zeta,\nu)$ is the matrix in \eqref{eq:kkt}, which is nonsingular because $\mu I\succ0$ and linear independence ensures that $M$ has full row rank.

Step 2 gives the sensitivity with respect to a right-hand-side entry.
Differentiating the two KKT equalities with respect to a single entry $b_i$ yields $\mu\,\partial\zeta+M^\xT\partial\nu=0$ and $M\,\partial\zeta=e_i$, where $e_i\in\real^{r_M}$ is the unit vector of row $i$. Let $(u,p)$ denote the solution to the adjoint system \eqref{eq:kkt}. The first block row of this system is $\mu u+M^\xT p=\dsenspad$, and the second is $Mu=0$. Substituting these expressions into the quantity of interest yields
\begin{equation*}
\makebox[0pt]{$\displaystyle \dsenspad^\xT\partial\zeta=(\mu u+M^\xT p)^\xT\partial\zeta=u^\xT(\mu\,\partial\zeta)+p^\xT(M\,\partial\zeta)=-u^\xT M^\xT\partial\nu+p_i=-(Mu)^\xT\partial\nu+p_i=p_i,$}
\end{equation*}
which is the first formula in \eqref{eq:adjoint}.

Step 3 gives the sensitivity with respect to a matrix entry.
Differentiating with respect to an entry $A_{ij}$ of an active row introduces the term $(\partial M^\xT)\nu$ into the stationarity equality, where the only nonzero coordinate is $\nu_i$ in position $j$. It also introduces the term $(\partial M)\zeta$ into the feasibility equality, where the only nonzero coordinate is $\zeta_j$ in position $i$. Applying the same substitution yields
\begin{equation*}
\makebox[0pt]{$\displaystyle \dsenspad^\xT\partial\zeta=u^\xT\big(-M^\xT\partial\nu-(\partial M^\xT)\nu\big)+p^\xT\big(-(\partial M)\zeta\big)=-\nu_i u_j-\zeta_j p_i ,$}
\end{equation*}
which is the second formula in \eqref{eq:adjoint}.

Since $\dsens^\xT\hat x_\mu=\dsenspad^\xT\zeta^\star$, both formulas represent exact scenario sensitivities of the regularized relaxation.\Halmos
\end{proof}

\begin{remark}[On the degradation of the gradient]\label{rem:bias}
Outside the conditions specified in Proposition~\ref{prop:exact}, the gradient remains computable but is inexact in three distinct ways. First, when the recourse is integer, the duals in \eqref{eq:dualgrad} correspond to those of its linear relaxation, causing both factors to inherit the integrality gap as bias. The resulting direction is then an approximation, and neither its accuracy nor its magnitude is guaranteed. Second, when the recourse is a degenerate linear program, as is the case for transportation and assignment structures, the dual solution is not unique. Consequently, $\dsens$ depends on a quantity that is not well defined and may change discontinuously with the scenarios. Third, when the first stage is integer, the direction combines the gradient of $F$ at the integer decision, computed as if the first-stage variables were continuous, with the Jacobian of the relaxed map. Thus, it serves as a surrogate direction by construction, even when the recourse is a linear program with unique duals. All three phenomena are present in the benchmark, and together they motivate the development of the next subsection, which ensures that an inexact $\gDF$ lowers the sampling efficiency while the expected step still descends.
\end{remark}

\subsection{Guided Evolution Strategies with a Decision-Focused Subspace}\label{sec:guided}

This subsection develops an estimator in which the structural direction from Section~\ref{sec:dfgrad} influences only the sampling process. Section~\ref{sec:gestimator} defines the smoothed objective on which the search operates and constructs the guided estimator, Section~\ref{sec:descent} proves that the expected update is a descent direction for any choice of guiding subspace, and Sections~\ref{sec:subDF} and~\ref{sec:subH} construct the subspace $\Usub$ using two approaches. One offers higher accuracy but requires additional problem-specific information, while the other is computationally less expensive and relies solely on evaluated costs. Section~\ref{sec:algorithm} summarizes the algorithm.

\subsubsection{The Guided Estimator}\label{sec:gestimator}

The objective $\Fcomp(\Zbar)=F\big(\hat x(S'(\Zbar))\big)$ is piecewise constant in $\Zbar$. When a centroid is moved, the reduced problem changes continuously, but the optimal first-stage decision remains unchanged until a switch occurs. Consequently, the objective is flat for almost every $\Zbar$ and exhibits discontinuities at switching points. Its gradient is zero wherever it exists and undefined at the discontinuities, and a finite difference yields zero unless the perturbation crosses a discontinuity. Averaging the objective over a neighbourhood of $\Zbar$ recovers this information, as the average weights the values of $\Fcomp$ by the probability mass assigned by the neighbourhood. Thus, moving $\Zbar$ shifts mass from one value to another, and the average acquires a gradient wherever a discontinuity is within reach. The natural smooth surrogate is therefore the Gaussian-smoothed objective. For a smoothing radius $\sigma>0$ and a symmetric positive definite covariance $\Sigmacov\in\real^{d\times d}$, where $d=mK$ and the centroids $\Zbar$ are read as a vector in $\real^{d}$, define
\begin{equation}\label{eq:smooth}
\Ftilde_{\sigma}(\Zbar)\;=\;\expect_{\varepsilon\sim\mathcal N(0,\Sigmacov)}\big[\,\Fcomp(\Zbar+\sigma\varepsilon)\,\big],
\end{equation}
which is differentiable for every $\sigma>0$, despite the non-differentiability of the underlying objective.

Guided Evolutionary Strategies \citep{maheswaranathan2019guided} selects the covariance as a combination of an isotropic floor and a low-dimensional informed subspace,
\begin{equation}\label{eq:sigma}
\Sigmacov=\frac{\alpha}{d}\,I_d+\frac{1-\alpha}{\ell}\,\Usub\Usub^\xT,
\qquad \operatorname{tr}\Sigmacov=1,\quad \alpha\in(0,1],
\end{equation}
where the columns of $\Usub\in\real^{d\times\ell}$ form an orthonormal basis of the guiding subspace, satisfying $\Usub^\xT\Usub=I_\ell$. Here, $\ell$ denotes the number of guiding directions, and $d=mK$ represents the number of latent coordinates along which the search proceeds, with $\ell\ll d$. The unit trace fixes the total variance of the distribution, so $\alpha$ redistributes this variance between the isotropic floor and the subspace without altering the overall scale of the search.

Each probe is generated as $\varepsilon=\sqrt{\alpha/d}\,\xi_d+\sqrt{(1-\alpha)/\ell}\,\Usub\xi_\ell\in\real^{d}$, where $\xi_d\sim\mathcal N(0,I_d)$ and $\xi_\ell\sim\mathcal N(0,I_\ell)$ are independent, and the resulting covariance is $\Sigmacov$. A probe is called antithetic when it is evaluated at the two opposite points $\Zbar+\sigma\varepsilon$ and $\Zbar-\sigma\varepsilon$, a standard variance-reduction scheme known as antithetic sampling \citep{maheswaranathan2019guided}. Drawing $P$ antithetic probes $\varepsilon_1,\dots,\varepsilon_P\sim\mathcal N(0,\Sigmacov)$ yields the estimator
\begin{equation}\label{eq:gestimator}
\widehat g_{\Zbar}
=\frac{1}{2\sigma P}\sum_{i=1}^{P}
\big(\Fcomp(\Zbar+\sigma\varepsilon_i)-\Fcomp(\Zbar-\sigma\varepsilon_i)\big)\,\varepsilon_i
\end{equation}
The SPSA estimator of \name{DGSS} is the special case with a single probe, $P=1$, and a Rademacher direction $\Delta$ in place of $\varepsilon_i$, that is, $\tfrac{1}{2\sigma}\big(\Fcomp(\Zbar+\sigma\Delta)-\Fcomp(\Zbar-\sigma\Delta)\big)\Delta$. The estimator \eqref{eq:gestimator} differs from it by averaging $P$ probes and by drawing them from $\mathcal N(0,\Sigmacov)$, whose covariance is shaped by the guiding subspace. The estimate $\widehat g_{\Zbar}\in\real^{d}$ is used as the gradient of the objective, so the centroids move in the direction $-\widehat g_{\Zbar}$, which is determined entirely by the evaluated objective values of the probes. Consequently, the guide influences the estimator solely through the distribution from which these probes are sampled.

\subsubsection{Descent Guarantee}\label{sec:descent}

The guarantees provided below pertain to the smoothed objective $\Ftilde_\sigma$ as defined at the current iteration, since $\sigma$ and $\Usub$ vary between iterations and the smoothed objective is updated accordingly. The assumptions required by these guarantees concern $\Fcomp$ itself.

\begin{lemma}[Smoothed-gradient identity, \citealt{nesterov2017random}]\label{lem:smooth}
Assume that the objective $\Fcomp:\real^d\to\real$ is measurable and has at most polynomial growth. Let $\sigma>0$ and $\Sigmacov\succ0$. Under these conditions, $\Ftilde_\sigma$ in \eqref{eq:smooth} is continuously differentiable. Furthermore, the single-probe antithetic estimator $\widehat g^{(1)}=\tfrac{1}{2\sigma}\big(\Fcomp(\Zbar+\sigma\varepsilon)-\Fcomp(\Zbar-\sigma\varepsilon)\big)\varepsilon$ with $\varepsilon\sim\mathcal N(0,\Sigmacov)$ satisfies
\begin{equation}\label{eq:lemma}
\expect\big[\widehat g^{(1)}\big]\;=\;\Sigmacov\,\nabla\Ftilde_\sigma(\Zbar)
\end{equation}
\end{lemma}

\begin{proof}{Proof of Lemma~\ref{lem:smooth}.}
Under the change of variables $\tau=\Zbar+\sigma\varepsilon$, the smoothed objective \eqref{eq:smooth} becomes
\begin{equation}\label{eq:app_shift}
\Ftilde_\sigma(\Zbar)=\int_{\real^d} \Fcomp(\tau)\,\rho(\tau\mid\Zbar)\,d\tau,
\end{equation}
where $\rho(\cdot\mid\Zbar)$ is the density of the Gaussian distribution $\mathcal N(\Zbar,\sigma^2\Sigmacov)$,
\begin{equation*}
\makebox[0pt]{$\displaystyle \rho(\tau\mid\Zbar)=\big(2\pi\sigma^2\big)^{-d/2}|\Sigmacov|^{-1/2}\exp\!\Big(-\tfrac{1}{2\sigma^{2}}\,(\tau-\Zbar)^\xT\Sigmacov^{-1}(\tau-\Zbar)\Big)$}
\end{equation*} Since $\Fcomp$ is measurable with at most polynomial growth, and the Gaussian density along with all its derivatives in $\Zbar$ decays faster than any polynomial, uniformly for $\Zbar$ in a bounded set, the integrand and its derivatives in $\Zbar$ are bounded by an integrable function of $\tau$. The derivative can therefore be taken inside the integral by the Leibniz rule for differentiation under the integral sign \citep[Theorem~2.27]{folland1999real}, which follows from the dominated convergence theorem. Consequently, $\Ftilde_\sigma$ inherits the smoothness of the density. Applying $\nabla_{\Zbar}\rho = \rho\,\nabla_{\Zbar}\log \rho$ and
\begin{equation}\label{eq:app_score}
\nabla_{\Zbar}\log \rho(\tau\mid\Zbar)=\frac{\Sigmacov^{-1}(\tau-\Zbar)}{\sigma^{2}},
\end{equation}
and substituting back $\tau=\Zbar+\sigma\varepsilon$ yields the identity
\begin{equation}\label{eq:app_onesided}
\nabla\Ftilde_\sigma(\Zbar)=\expect\Big[\Fcomp(\Zbar+\sigma\varepsilon)\,\frac{\Sigmacov^{-1}\varepsilon}{\sigma}\Big]
\end{equation}
Multiplying both sides of \eqref{eq:app_onesided} by $\sigma\Sigmacov$ gives the equivalent expression $\expect\big[\Fcomp(\Zbar+\sigma\varepsilon)\,\varepsilon\big]=\sigma\,\Sigmacov\,\nabla\Ftilde_\sigma(\Zbar)$. Since $\varepsilon$ and $-\varepsilon$ are identically distributed under $\mathcal N(0,\Sigmacov)$,
$\expect[\Fcomp(\Zbar-\sigma\varepsilon)\,\varepsilon]=-\expect[\Fcomp(\Zbar+\sigma\varepsilon)\,\varepsilon]$, so the antithetic difference doubles the one-sided term,
\begin{equation}\label{eq:app_anti}
\expect\Big[\big(\Fcomp(\Zbar+\sigma\varepsilon)-\Fcomp(\Zbar-\sigma\varepsilon)\big)\,\varepsilon\Big]
=2\sigma\,\Sigmacov\,\nabla\Ftilde_\sigma(\Zbar)
\end{equation}
Dividing by $2\sigma$ yields $\expect[\widehat g^{(1)}]=\Sigmacov\nabla\Ftilde_\sigma(\Zbar)$, as shown in \eqref{eq:lemma}.\Halmos
\end{proof}

The assumptions of the lemma hold for the objective of this paper. The decision returned by the reduced problem ranges over a bounded set on which $Q(x,s)$ is finite, so $\Fcomp$ is bounded, which satisfies the growth condition. Furthermore, $\Fcomp$ is measurable, since it is piecewise constant when the first stage is integer and piecewise smooth otherwise. Continuity of $\Fcomp$ is not required, since differentiability is established for the smoothed objective $\Ftilde_\sigma$ and not assumed for $\Fcomp$. The lemma therefore covers objectives that are discontinuous where the reduced problem changes its integer decision.

The antithetic form is typically preferred in practice because it eliminates the zeroth-order term $\expect[\Fcomp(\Zbar)\varepsilon]=0$, thereby reducing variance without introducing bias. Averaging over $P$ independent probes reduces the variance by a factor of $P$ while leaving the expectation unchanged. It is important to note that the estimator is unbiased for $\Sigmacov\nabla\Ftilde_\sigma$, which is the gradient of the smoothed objective preconditioned by the covariance. In the isotropic case $\alpha=1$ the covariance is $\Sigmacov=I_d/d$ and the expectation is $\nabla\Ftilde_\sigma(\Zbar)/d$, a positive multiple of the smoothed gradient. For $\alpha<1$ the expectation is tilted toward the guiding subspace. The estimator is not unbiased for $\nabla\Fcomp$ in general, and for the piecewise-constant objective of this paper it never is. When $\Fcomp$ is smooth, $\nabla\Ftilde_\sigma$ differs from $\nabla\Fcomp$ by a term of order $\mathcal O(\sigma^{2})$.

The following proposition and its proof are based on \citet{maheswaranathan2019guided}, who assume a differentiable objective. The extension here is that Lemma~\ref{lem:smooth} requires only measurability.

\begin{proposition}[Descent for every subspace]\label{prop:descent}
Under the assumptions of Lemma~\ref{lem:smooth}, let $\Sigmacov$ be defined as in \eqref{eq:sigma} with $\alpha\in(0,1]$ and any orthonormal $\Usub$. The estimator \eqref{eq:gestimator} then satisfies $\expect[\widehat g_{\Zbar}]=\Sigmacov\nabla\Ftilde_\sigma(\Zbar)$, and whenever $\nabla\Ftilde_\sigma(\Zbar)\neq0$,
\begin{equation}\label{eq:descentbound}
\makebox[0pt]{$\displaystyle \big\langle \nabla\Ftilde_\sigma(\Zbar),\, \expect[\widehat g_{\Zbar}]\big\rangle=\nabla\Ftilde_\sigma(\Zbar)^\xT\,\Sigmacov\,\nabla\Ftilde_\sigma(\Zbar)\ge\frac{\alpha}{d}\,\big\|\nabla\Ftilde_\sigma(\Zbar)\big\|^2\;>\;0$}
\end{equation}
Hence $-\expect[\widehat g_{\Zbar}]$ is a descent direction for $\Ftilde_\sigma$ at $\Zbar$, for every choice of the guiding subspace $\Usub$.
\end{proposition}

\begin{proof}{Proof.}
The identity $\expect[\widehat g_{\Zbar}]=\Sigmacov\nabla\Ftilde_\sigma(\Zbar)$ is established by Lemma~\ref{lem:smooth} and the linearity of \eqref{eq:gestimator} with respect to its $P$ independent probes. Given that $\Usub\Usub^\xT\succeq0$, the covariance \eqref{eq:sigma} satisfies $\Sigmacov\succeq(\alpha/d)I_d$, and \eqref{eq:descentbound} follows as the corresponding quadratic-form bound.\Halmos
\end{proof}

\begin{remark}[What the guide changes]\label{rem:variance}
Proposition~\ref{prop:descent} determines the sign of the expected step for every subspace, whereas the subspace determines the variance of a single step. To quantify this effect, consider an objective that is linear in the vicinity of the centroids, and measure the quality of a single antithetic probe by the ratio of the squared norm of its mean to the trace of its covariance, which acts as a signal-to-noise ratio \citep{maheswaranathan2019guided}. For $d=64$ and $\alpha=0.01$, this ratio is approximately $0.50$ when the guiding direction is aligned with the gradient, approximately $0.015$ under isotropic sampling, and approximately $0.00016$ when the guiding direction is orthogonal to the gradient. An aligned guide is therefore approximately $32$ times more efficient than isotropic sampling, whereas a misaligned guide is approximately $100$ times less efficient. An inexact guide consequently reduces the sampling efficiency without affecting the sign of the expected step. The mixture parameter $\alpha$ governs this trade-off. \citet{maheswaranathan2019guided} show that the optimal $\alpha$ approaches zero as the alignment improves and approaches one as it deteriorates. The selected $\alpha$ therefore reflects the quality of the subspace on the families where the gap depends on it, as confirmed by the computational study.
\end{remark}

\begin{remark}[Scope of the guarantee]\label{rem:scope}
Proposition~\ref{prop:descent} pertains to the estimate at the centroids. Algorithm~\ref{alg:gdgss} propagates this estimate back to the encoder parameters via the Jacobian of the centroid map and applies it using the autoencoder optimizer. The sign guarantee does not extend through this propagation, which is typical for evolution strategies applied through a network. The estimator in \eqref{eq:gestimator} follows \citet{maheswaranathan2019guided} with their scale $\beta$ set to one, and at $\alpha=1$ it reduces to isotropic evolution strategies.
\end{remark}

\begin{figure}[t]
\FIGURE
{\includegraphics[width=0.75\textwidth]{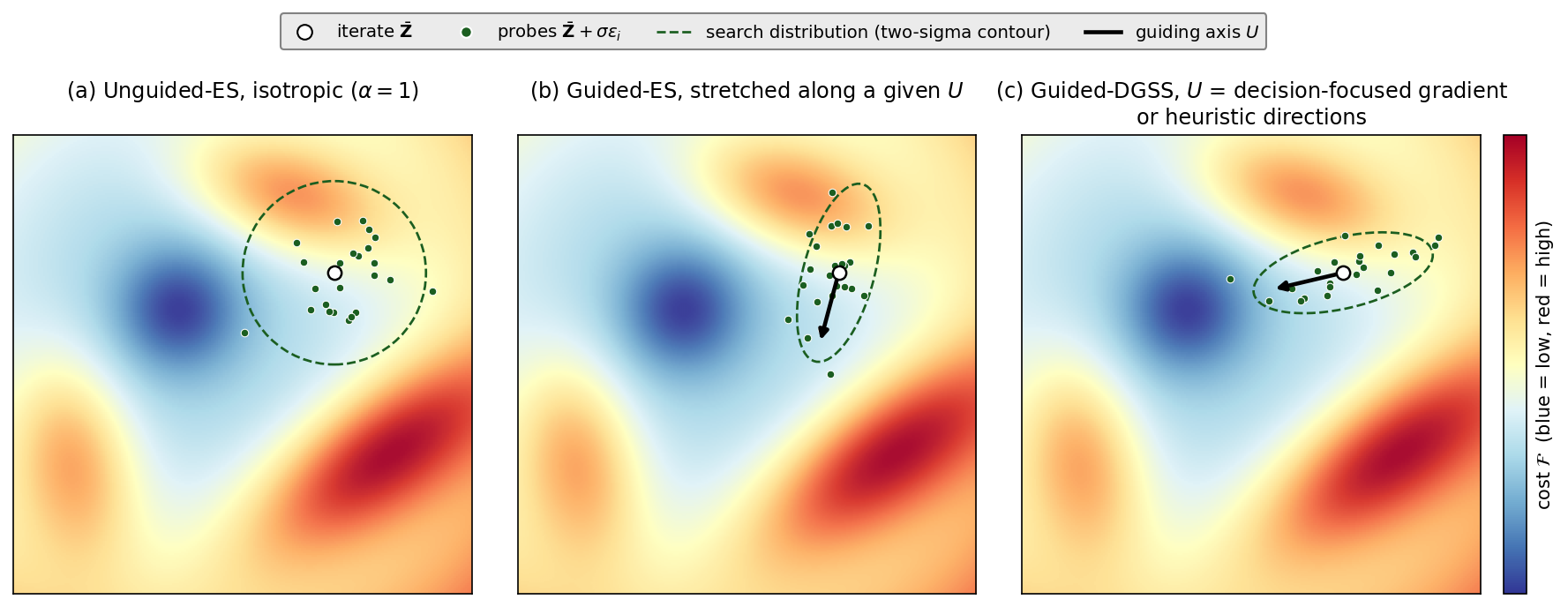}}
{Three search distributions at a single iteration on the cost surface of Figure~\ref{fig:estimators}. Each panel displays $P=24$ probes sampled from the covariance \eqref{eq:sigma}, with the dashed ellipse indicating the two-sigma contour. For $\alpha=1$ (left), the search distribution is isotropic. Guided Evolutionary Strategies (middle) elongates the distribution along a generic axis $U$, while \gdgss\ (right) aligns $U$ with either the decision-focused gradient from Section~\ref{sec:subDF} or the heuristic directions from Section~\ref{sec:subH}, orienting the stretch toward the low-cost region. The isotropic component ensures coverage in all directions, so even if $U$ is misaligned, descent in expectation is maintained. The surface is drawn smooth for clarity. In the setting of the paper it is piecewise constant and only the smoothed objective possesses a gradient.\label{fig:tilt}}
{}
\end{figure}

Figure~\ref{fig:tilt} illustrates the geometry. The guiding subspace stretches the search distribution in the directions indicated by the decision-focused gradient, whereas the isotropic floor ensures that every direction remains covered.

\subsubsection{The Decision-Focused Subspace (\guidedsurr)}\label{sec:subDF}
\guidedsurr\ constructs the subspace using the gradient described in Section~\ref{sec:dfgrad}, assigning $\Usub$ as the current decision-focused direction $\gDF_{\Zbar}$ normalized to unit length. Thus, $\ell=1$ and the search distribution is elongated along this single direction. This direction represents the exact decision sensitivity under the conditions specified in Proposition~\ref{prop:exact}, and serves as a relaxation-based approximation otherwise. When derived from the linear relaxation of an integer reduced problem, it inherits the bias noted in Remark~\ref{rem:bias}. The isotropic component of the covariance \eqref{eq:sigma} ensures that all other directions remain accessible, and the computational study selects a larger $\alpha$ on most of the instances where the direction is only approximate.

\subsubsection{The Heuristic Fallback Subspace (\guidedes)}\label{sec:subH}
The decision-focused subspace requires two prerequisites, recourse duals and a differentiable regularized solve. Remark~\ref{rem:bias} demonstrates that the quality of this subspace diminishes precisely when these prerequisites are weakened, such as under integrality constraints or degenerate duals. To address this limitation, a fallback guide, \guidedes, is proposed. This guide does not require duals or a differentiable solve, instead relying solely on per-scenario costs computed by the loop, and thus remains informative in situations where the decision-focused guide is unreliable. Proposition~\ref{prop:descent} ensures descent for every subspace, so incorporating this more economical guide does not compromise validity and provides practitioners with an alternative when the reduced problem must be treated as a black box. Each direction constructed by this guide follows a consistent procedure. Empirical scenarios are scored, a target scenario $s^\star$ is selected, and the nearest centroid is moved toward it. Each empirical scenario $s$ is assigned a nonnegative score $\varphi_s\in\real_{+}$, and $s^\star$ is the scenario with the highest score among those not already targeted by an earlier score in the same iteration, so that the four directions aim at four different scenarios. Let $\mathbf z_s=E_{\theta_e}(\mathbf w_s)\in\real^{m}$ denote the latent code of scenario $s$ and $k^\star=\arg\min_k\|\bar{\mathbf z}_k-\mathbf z_{s^\star}\|_2$ the centroid nearest to the target. The direction is defined as the matrix that is zero except in column $k^\star$, where it equals the unit vector $(\mathbf z_{s^\star}-\bar{\mathbf z}_{k^\star})/\|\mathbf z_{s^\star}-\bar{\mathbf z}_{k^\star}\|_2$. Perturbing the centroids along this direction moves centroid $\bar{\mathbf z}_{k^\star}$ toward $\mathbf z_{s^\star}$, so that the synthetic scenario $\hat{\mathbf w}'_{k^\star}$ decoded from it represents the target scenario $s^\star$ better. Let $Q_s=Q(\hat x,s)\in\real$ denote the recourse cost of scenario $s$ under the current decision $\hat x$, which is obtained as a by-product of the full-set evaluation and requires no additional solves. The four scores are as follows.
\begin{itemize}
\item \textbf{Residual} ($\varphi_s=Q_s$). The score corresponds to the recourse cost itself, identifying the scenario that the current decision serves at the highest expense. This is the most direct cost signal the solver provides.
\item \textbf{Coverage} ($\varphi_s=\min_k\|\mathbf z_s-\bar{\mathbf z}_k\|_2^2$). The score measures the latent-space distance to the nearest centroid, identifying the scenario that the current reduced set represents least accurately. This criterion is purely geometric, does not require a cost signal, and is therefore always available.
\item \textbf{Deviation} ($\varphi_s=\max\{Q_s-\bar Q_{\kappa(s)},0\}$, where $\kappa(s)$ denotes the centroid scenario $s$ is assigned to and $\bar Q_k$ is the mean cost of cluster $k$). The score quantifies the extent to which a scenario's cost exceeds the average cost of its assigned cluster. This approach targets scenarios that are expensive relative to the other scenarios of their cluster, even if their absolute cost is moderate.
\item \textbf{Tail risk} (CVaR). Candidates are limited to the $\lceil\eta|S|\rceil$ scenarios with the highest costs, where $\eta=0.1$ ensures that only the worst decile is retained. The target scenario is the highest-cost scenario in this subset that is not already targeted. This strategy focuses on expensive scenarios, thereby improving the reduced set's representation of worst-case recourse.
\end{itemize}
All four directions are stacked as columns and orthonormalized to form $\Usub$. Isotropic exploration is excluded from the subspace because the $(\alpha/d)I_d$ term in \eqref{eq:sigma} already accounts for it.

\begin{algorithm}[t]
\caption{\gdgss}
\label{alg:gdgss}
\renewcommand{\algorithmicrequire}{\textbf{Input:}}
\renewcommand{\algorithmicensure}{\textbf{Output:}}
\begin{algorithmic}[1]
\Require empirical scenarios $S$, reduced-set size $K$, covariance mix $\alpha$, number of probes $P$, time budget
\Statex \textbf{Step \stepa\ (pretraining)}
\State Train the autoencoder $(E_{\theta_e},D_{\theta_d})$ on $S$ by minimizing $\Lmse$
\Statex \textbf{Step \stepb\ (search)}
\Repeat
\State Encode $S$, form differentiable centroids $\Zbar$, decode $S'(\Zbar)$ by \eqref{eq:decode}
\State Solve the reduced problem \eqref{eq:reduced} and evaluate $\Lobj=F(\hat x)$ on $S$ by \eqref{eq:obj_eval}
\State Build $\Usub$ from $\gDF_{\Zbar}$ via \eqref{eq:dualgrad}--\eqref{eq:adjoint} (\guidedsurr) or from heuristic scores (\guidedes)
\State Set $\sigma$ to the empirical standard deviation of the entries of $\Zbar$, draw antithetic probes with the covariance \eqref{eq:sigma} and compute $\widehat g_{\Zbar}$ by \eqref{eq:gestimator}
\State Inject $\widehat g_{\Zbar}$ at $\Zbar$ as the gradient of $\Lobj$ and update $\theta$ by backpropagation of $\Lmse+\Lobj$
\Until{the time budget is reached}
\Ensure the evaluated decision with the lowest full-set objective and the synthetic scenarios that induced it
\end{algorithmic}
\end{algorithm}

\subsubsection{The \gdgss\ Algorithm}\label{sec:algorithm}
Algorithm~\ref{alg:gdgss} provides an overview of the method. The computational cost per iteration is primarily determined by the $2P$ full-set evaluations of the antithetic probes, which are mutually independent and thus amenable to parallel computation. The decision-focused guide adds a dual pass, which under integer recourse requires one linear solve per scenario, one solve of the regularized relaxation, and one sparse solve of \eqref{eq:kkt}. Consistent with \name{DGSS}, the perturbation radius $\sigma$ is updated at each iteration to match the empirical standard deviation of the entries of $\Zbar$, ensuring that the perturbation scale reflects the dispersion of the centroids in the latent space. Compared to \name{DGSS}, the only modification is in the gradient estimate specified in \eqref{eq:gz}.

\section{Computational Study}\label{sec:computational}

This section evaluates the proposed estimators on six families of two-stage stochastic programs and addresses three primary questions. The first question is whether the guided estimators identify superior reduced sets compared to \plainm\ under identical conditions and equivalent wall-clock budgets. The second question examines whether the observed improvement is attributable to the guiding subspace or merely to the increased number of probes. This is assessed using an unguided estimator that employs the same probes (\isoes). The third question investigates the utility of the decision-focused direction when it is applied directly as the update, rather than as a means of shaping the sampling process. This is tested with a deterministic estimator that follows the direction without probes (\analytic). Additionally, the reduced sets are compared with a two-hour solution of the full problem by an exact solver, in contrast to $20$ minutes allocated for each search, to quantify the time savings achieved through reduction and to demonstrate that, within a limited timeframe, a reduced set can yield a better decision than the exact solver achieves using all scenarios.

Table~\ref{tab:methods} lists the five methods. Section~\ref{sec:setup} outlines the benchmark and protocol, Section~\ref{sec:exact} compares the reduced sets to the exact solver, Section~\ref{sec:aggregate} presents results both pooled and by family, identifies the source of performance gains, and connects these to the recourse structure. Section~\ref{sec:budget} examines the gap over wall-clock time, while Section~\ref{sec:scenarios} analyzes the synthetic scenario sets.

\begin{table}[t]
\TABLE
{The five methods of the study. All use the \name{DGSS} loop of Algorithm~\ref{alg:gdgss} and differ only in the gradient estimate. Probes are full-set evaluations per iteration.\label{tab:methods}}
{\begin{adjustbox}{max width=\textwidth}
\begin{tabular}{l l c l}
\toprule
Method & Gradient estimate & Probes & Role in the study \\
\midrule
\plainm & SPSA along one random direction (Section~\ref{sec:pipeline}) & $2$ & baseline, the estimator of \name{DGSS} \\
\analytic & decision-focused gradient (Section~\ref{sec:dfgrad}) used directly as the update & $0$ & control, tests the effect of the direction without sampling \\
\isoes & evolution strategy with isotropic covariance, $\alpha=1$ & $16$ & control, tests the effect of the probes without a guide \\
\guidedsurr & guided evolution strategy, decision-focused subspace, $\ell=1$ (Section~\ref{sec:subDF}) & $16$ & proposed, guided by the decision-focused subspace \\
\guidedes & guided evolution strategy, heuristic subspace, $\ell=4$ (Section~\ref{sec:subH}) & $16$ & proposed, guided by heuristic directions \\
\bottomrule
\end{tabular}
\end{adjustbox}}
{}
\end{table}

\subsection{Benchmark and Protocol}\label{sec:setup}

The benchmark consists of six problem families, selected to ensure that the recourse structures described in Proposition~\ref{prop:exact} and Remark~\ref{rem:bias} are all represented. Additionally, for at least some families, the full problem is computationally intractable for an exact solver within a practical timeframe. Table~\ref{tab:benchmark} enumerates these families, while Section~\ref{ec:models} of the e-companion details the model specifications, distributions, and parameter ranges used to generate instances. CFLP refers to the capacitated facility location problem as defined by \citet{pisinger2026deep}, following the framework of \citet{wu2022learning,wu2024hgcn2sp}, with binary opening decisions and binary customer assignments in the recourse. DCAP denotes a dynamic capacity acquisition and allocation problem, structured according to the DCAP instances in \citet{ahmed2003dynamic} from the SIPLIB test library \citep{ahmed2015siplib}, but generated at a larger scale. This problem features continuous capacity expansions and binary expansion decisions in the first stage, with binary client assignments in the recourse. ELECTRICITY is a problem from the public Cornell stochastic-programming test set, representing a power capacity planning model in the family of \citet{louveaux1988optimal}. This model is a linear program in both stages, so the conditions of Proposition~\ref{prop:exact} are satisfied except in cases of dual degeneracy in the transportation recourse, and the gradient is then exact up to the effect of the proximal term. EV is the electric-vehicle charging-station location problem from \citet{pisinger2026deep}, characterized by an integer first stage and a linear transportation recourse. SUC is a stochastic unit commitment problem with 200 generators and 24 periods, featuring a binary first stage and a linear dispatch recourse in which randomness is introduced through equality constraints. VRPSD is a vehicle routing problem with stochastic demands, involving binary routing decisions in the first stage and binary delivery decisions in the recourse, and the random demands affect the recourse as coefficients in the vehicle capacity constraints. Table~\ref{tab:benchmark} indicates, for each family, whether randomness enters the right-hand side or the constraint matrix of the recourse. Each family is evaluated at a single problem size with three instances. Every instance includes $|S|=200$ empirical scenarios. Each method is tested for reduced-set sizes $K\in\{1,2,3\}$ with three training seeds, resulting in $27$ runs per family and $162$ runs per method. The extensive forms range from $8{,}460$ columns for VRPSD to $7.9$ million columns for EV.

\begin{table}[t]
\TABLE
{The six test families, with their models and instance generation procedures described in Section~\ref{ec:models} of the e-companion. Variable types are specified for both the first stage and the recourse, where mixed binary refers to the presence of both binary and continuous variables. The column Randomness in indicates whether the random entries of a scenario affect the right-hand side (RHS) or the constraint matrix of the recourse. The scenario dimension denotes the length of the scenario vector, which also corresponds to the autoencoder input size. The extensive form represents the deterministic equivalent across all $200$ scenarios, with the associated counts of rows, columns, and integer variables.\label{tab:benchmark}}
{\begin{adjustbox}{max width=\textwidth}
\begin{tabular}{l l ll cc r rrr}
\toprule
\multirow{2}{*}{Family} & \multirow{2}{*}{Problem} & \multicolumn{2}{c}{Variables} & \multicolumn{2}{c}{Randomness in} & Scenario & \multicolumn{3}{c}{Extensive form, 200 scenarios} \\
\cmidrule(lr){3-4}\cmidrule(lr){5-6}\cmidrule(lr){8-10}
 & & first stage & recourse & RHS & matrix & dimension & rows & columns & integer \\
\midrule
CFLP & capacitated facility location & binary & mixed binary & $\checkmark$ & $\checkmark$ & 840 & 16{,}001 & 164{,}020 & 160{,}020 \\
DCAP & dynamic capacity acquisition & mixed binary & binary &  & $\checkmark$ & 360 & 19{,}236 & 84{,}072 & 84{,}036 \\
ELECTRICITY & power capacity planning & continuous & continuous & $\checkmark$ &  & 150 & 40{,}025 & 300{,}050 & 0 \\
EV & charging station location & integer & continuous & $\checkmark$ &  & 960 & 384{,}000 & 7{,}872{,}040 & 40 \\
SUC & unit commitment & binary & continuous & $\checkmark$ &  & 24 & 3{,}774{,}400 & 974{,}400 & 9{,}600 \\
VRPSD & vehicle routing, stochastic demand & mixed binary & binary &  & $\checkmark$ & 30 & 8{,}786 & 8{,}460 & 8{,}360 \\
\bottomrule
\end{tabular}
\end{adjustbox}}
{}
\end{table}

\begin{table}[t]
\TABLE
{Results of the exact solver on the extensive form of each instance, with all $200$ scenarios and a two-hour time limit. The status column indicates whether the solver proved optimality or terminated upon reaching the time limit. Solve time refers to the wall-clock time, with the two-hour limit displayed for instances that did not complete within the allotted period. The gap represents the difference between the final incumbent and the final lower bound, relative to the bound, which serves as the metric for the methods' gap to bound. The final two columns report the time at which the solver completed the root relaxation and when it first attained its final incumbent.\label{tab:exactref}}
{\begin{adjustbox}{max width=\textwidth}
\begin{tabular}{l c l r rr r r r}
\toprule
Family & Instance & Status & Solve time & Final incumbent & Lower bound & Gap (\%) & Root relaxation solved & Final incumbent found \\
\midrule
CFLP & 1 & time limit & 120 min & 8{,}119.77 & 7{,}629.24 & 6.43 & 8 s & 113 min \\
 & 2 & time limit & 120 min & 7{,}908.57 & 7{,}906.04 & 0.03 & 8 s & 112 min \\
 & 3 & time limit & 120 min & 8{,}894.40 & 7{,}564.07 & 17.59 & 15 s & 120 min \\
\addlinespace[2pt]
DCAP & 1 & time limit & 120 min & 433.51 & 423.90 & 2.27 & 19 s & 88 min \\
 & 2 & time limit & 120 min & 436.89 & 424.30 & 2.97 & 18 s & 112 min \\
 & 3 & time limit & 120 min & 433.58 & 425.28 & 1.95 & 17 s & 110 min \\
\addlinespace[2pt]
ELECTRICITY & 1 & optimal & 2 s & -1{,}865.54 & -1{,}865.54 & 0 & 2 s & 2 s \\
 & 2 & optimal & 2 s & -1{,}832.81 & -1{,}832.81 & 0 & 2 s & 2 s \\
 & 3 & optimal & 2 s & -1{,}871.26 & -1{,}871.26 & 0 & 2 s & 2 s \\
\addlinespace[2pt]
EV & 1 & optimal & 49 min & 29{,}935{,}848 & 29{,}935{,}848 & 0 & 2.0 min & 7.6 min \\
 & 2 & optimal & 11 min & 25{,}870{,}315 & 25{,}870{,}315 & 0 & 1.8 min & 2.1 min \\
 & 3 & optimal & 9.4 min & 25{,}254{,}487 & 25{,}254{,}487 & 0 & 2.2 min & 4.3 min \\
\addlinespace[2pt]
SUC & 1 & time limit & 120 min & 13{,}805{,}057 & 13{,}804{,}671 & 0.003 & 21 min & 114 min \\
 & 2 & optimal & 103 min & 13{,}011{,}028 & 13{,}011{,}028 & 0 & 63 min & 103 min \\
 & 3 & optimal & 79 min & 12{,}470{,}997 & 12{,}470{,}997 & 0 & 6.8 min & 79 min \\
\addlinespace[2pt]
VRPSD & 1 & time limit & 120 min & 2{,}929.47 & 2{,}898.84 & 1.06 & 3 s & 22 min \\
 & 2 & time limit & 120 min & 2{,}598.89 & 2{,}576.09 & 0.89 & 2 s & 12 min \\
 & 3 & time limit & 120 min & 3{,}188.45 & 3{,}157.01 & 1.00 & 3 s & 26 min \\
\bottomrule
\end{tabular}
\end{adjustbox}}
{The exact solver is Gurobi 12 with ten threads and a zero optimality tolerance.}
\end{table}

For each instance, the reference is obtained by solving its extensive form with all $200$ scenarios in Gurobi with ten threads, zero optimality tolerance, and a two-hour time limit. Table~\ref{tab:exactref} presents these results. ELECTRICITY, formulated as a linear program, is solved in two seconds. The EV problem is solved to optimality in $9$ to $49$ minutes. SUC is solved in $79$ and $103$ minutes on two instances, with a gap of $0.003\%$ remaining on the third at the time limit. CFLP, DCAP, and VRPSD all reach the time limit with a remaining gap, ranging from $0.03\%$ to $17.6\%$ of the bound on CFLP, $2.0\%$ to $3.0\%$ on DCAP, and approximately $1\%$ on VRPSD. Consequently, performance is measured against the best solution identified by the exact solver within its time limit.
\begin{equation}\label{eq:gap}
\mathrm{Gap}(\%)=100\cdot\frac{F(\hat x)-F^{\mathrm{ref}}}{|F^{\mathrm{ref}}|}
\end{equation}
Here, $F^{\mathrm{ref}}=F(x^{\mathrm{ref}})$ is recomputed using the same exact full-set evaluation applied to all methods, where $x^{\mathrm{ref}}$ denotes the first-stage component of the solver's final incumbent, so the comparison is independent of the incumbent's recourse variables. The extensive-form solution serves as the reference for both decision quality and the lower bound. However, it is not necessarily the most effective method for these problems, as decomposition methods for two-stage integer programs may identify incumbents more rapidly. The subsequent comparison is therefore conducted between the reduced sets and this reference under matched wall-clock time. For ELECTRICITY, EV, and SUC, the reference is either optimal or within $0.003\%$ of optimality, resulting in a nonnegative gap. For CFLP, DCAP, and VRPSD, the reference is the two-hour incumbent, and a negative gap indicates that the reduced problem yielded a better decision than the exact solver within two hours. The gap to the exact solver's final lower bound is also reported, providing an upper limit on the distance to the true optimum, since no feasible decision can achieve an objective below this bound.

Each run follows the two-step procedure of \citet{pisinger2026deep}. The autoencoder is pretrained once per instance, seed, and $K$ for $10{,}000$ epochs, and all methods initiate the search from this pretrained state, so the five methods within a run differ solely in their gradient estimation approach. Pretraining requires between $22$ and $300$ seconds per run depending on the family, so the total time to obtain a method's $20$-minute result is about $25$ minutes at most, which is one fifth of the solver's two-hour limit. The multi-probe methods employ $P=8$ antithetic probes as defined in \eqref{eq:gestimator}, and the radius $\sigma$ of the centroid perturbation is set to the empirical standard deviation of the entries of $\Zbar$ at each iteration, consistent with \name{DGSS}. \plainm\ perturbs along an unnormalized Rademacher direction with entries of plus or minus one, positioning its probes at a distance of $\sigma\sqrt d$ from the centroids, whereas the probes of the evolution strategies possess unit-trace covariance and are located at a distance of approximately $\sigma$. This distinction is part of the comparison between \isoes\ and \plainm. The decision-focused subspace retains only the current direction, so $\ell=1$ for \guidedsurr, whereas the heuristic subspace stacks four directions, resulting in $\ell=4$ for \guidedes. The estimate $\widehat g_{\Zbar}$, and $\gDF_{\Zbar}$ for \analytic, is provided to backpropagation as the gradient of $\Lobj$ with respect to the centroids with unit scale, so the effective update on the centroids is performed by Adam with a learning rate of $0.0005$ for all methods, and no method terminates before the wall-clock limit. The two terms of $\mathcal L$ have unit weights, and for all methods the estimate is rescaled to a Frobenius norm of at most $2$ before it enters backpropagation, which keeps the update stable on families with large objective values. The full-set evaluation used to score each candidate is exact and consistent across all methods and the reference. The reduced problem and the recourse subproblems are solved using Gurobi 12.0.0. The search is conducted on ten cores of one node with a wall-clock budget of $20$ minutes, each run is evaluated by the best full-set objective achieved by any iteration completed within the budget, and the exact solver operates on the same cluster with ten threads on one node. The budget is one sixth of the solver's limit and is sufficient for the slowest family to complete approximately ten guided iterations. It is kept short in order to compare the reduced sets identified by the methods within the same limited time, although all methods would continue to improve with additional time.

The covariance mix $\alpha$ for each guided estimator is selected per instance from the five values $\{0.01, 0.1, 0.3, 0.5, 0.9\}$, all of which are evaluated. The value yielding the lowest mean full-set objective across the nine runs of the instance is chosen and applied for every $K$ and seed of that instance in all subsequent results. The selected values are provided in Appendix~\ref{app:alpha}, and Section~\ref{ec:alpha} of the e-companion presents the gap for each value. Statistical significance is assessed using the two-sided paired Wilcoxon signed-rank test on runs matched by family, instance, $K$, and seed. Since the integer first stage frequently results in two methods producing the same decision, the analysis reports wins, ties, and losses rather than solely a win rate. All runs of a given instance share its data and reference, so the p-values treat the pairs as exchangeable within a comparison and are not adjusted for multiple comparisons. Values near $0.05$ are considered indicative, while the primary conclusions are based on comparisons with $p$ below $0.001$.

\subsection{Boiling Two Hundred Scenarios Down to One}\label{sec:exact}

\begin{figure}[p]
\FIGURE
{\includegraphics[width=0.87\textwidth]{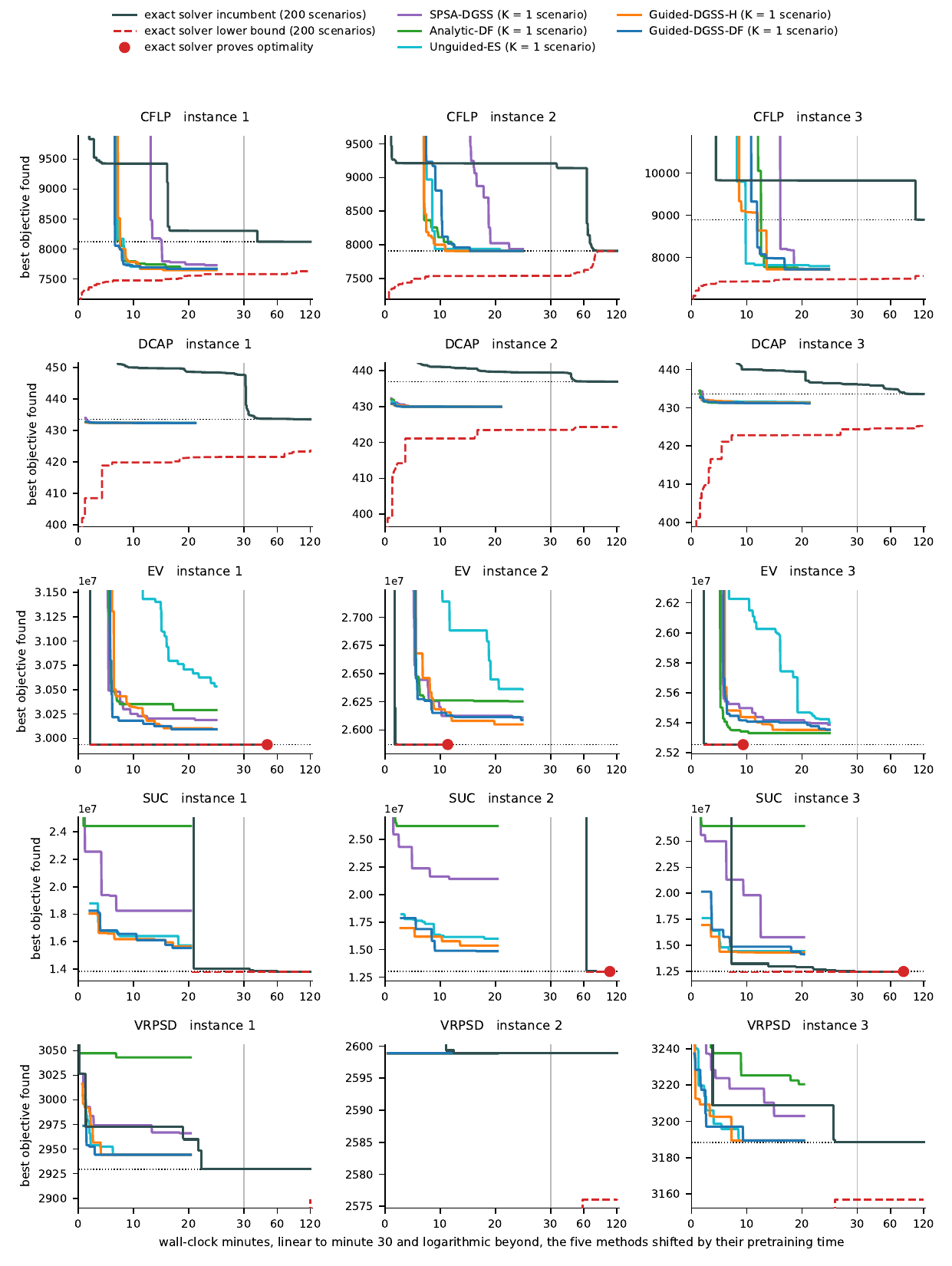}}
{Best full-set objective over wall-clock time at $K=1$, one panel per instance. For each method, the mean over its three training seeds is plotted against the exact solver's incumbent and lower bound on the $200$-scenario extensive form. The dotted horizontal line indicates the solver's final incumbent. The curves of the five methods end at their budget, pretraining plus $20$ minutes of search time, and the time axis ends at the solver's $2$-hour limit. A dot indicates where the solver proved optimality. ELECTRICITY is omitted because its extensive form is a linear program solved in two seconds. The five methods are shifted right by their pretraining time. The time axis is linear up to minute $30$ and logarithmic beyond it, as the grey vertical line marks. The $y$-axis range covers the curves after minute five of search, truncated at $20\%$ above the solver's final incumbent for CFLP and at $5\%$ for EV.\label{fig:exact}}
{}
\end{figure}

If a single synthetic scenario yields a decision of nearly the same quality as the full scenario set, the two-stage stochastic program can be replaced by a deterministic problem. This replacement has two advantages. The deterministic problem does not grow with the number of scenarios, and it can be solved with the exact and heuristic methods developed for deterministic combinatorial optimization, many of which have no direct counterpart for stochastic programs. This subsection shows that such a replacement is possible with a small loss in decision quality on most families. It compares the exact solver on the extensive form, with all $200$ scenarios and a two-hour time limit, to the reduction methods at $K=1$, and quantifies the loss due to this reduction. On CFLP and DCAP, the single scenario even yields a better decision in most runs than the exact solver finds on the full extensive form, in one fifth of the solver's time. The results for $K=1$ are shown in Figure~\ref{fig:exact}, and the results for $K=2$ and $K=3$ are provided in Section~\ref{ec:exactK} of the e-companion.

The MIP solver receives the complete problem and a time limit of two hours, while the five reduction methods are allotted $20$ minutes to identify the synthetic scenario that yields a strong first-stage decision. These methods access the $200$ scenarios only through candidate decision evaluation, which involves solving the $200$ recourse problems. Consequently, an iteration takes from less than one second on ELECTRICITY to about two minutes on SUC. The methods complete between ten iterations on SUC and about four thousand on ELECTRICITY, whereas the exact solver explores between one thousand and nine million nodes. Figure~\ref{fig:exact} presents the best full-set objective of each method over wall-clock time, averaged across three training seeds and shifted right by the pretraining time, so that the methods and the solver are both read from the moment the computation started. The figure is shown alongside the solver's incumbent and lower bound over two hours. Table~\ref{tab:instance} in Section~\ref{ec:instance} of the e-companion reports the final gap of each method to the solver's two-hour solution and to its lower bound for every instance and each value of $K$.

The outcome is determined by the size of the full problem, with the problem families divided into two groups. For CFLP and DCAP, the extensive forms contain $160{,}020$ and $84{,}036$ integer variables, respectively, and two hours of computation by the MIP solver are insufficient to solve them (Table~\ref{tab:exactref}). In these two families, the reduced sets compress the $200$ scenarios into one, are identified in one-fifth of the solver's time including pretraining, and in most runs yield a better decision than the solver achieves in two hours. For DCAP, each of the five reduction methods produces a better decision than the solver's two-hour solution in every run, and the solver never matches the $20$-minute result of \guidedsurr\ within its time limit.

For CFLP, the reduced set of \guidedsurr\ yields a better decision than the solver in $6$ of $9$ runs at $K=1$ and the same decision in the remaining $3$. Across all three values of $K$, it yields a better decision in $18$ of $27$ runs, the same decision in $8$, and a worse one in $1$. Superior decisions are observed in instances 1 and 3, where each of the five reduction methods outperforms the solver in every run except for a single run of \plainm\ on instance 3. Equal decisions occur in instance 2, where the solver comes within $0.03\%$ of its lower bound and the decision induced by the reduced set opens exactly the same facilities as the solver's solution. For CFLP, the margin is substantial. \guidedsurr\ achieves results $5.6\%$ below the solver's two-hour solution on instance 1 and $13.1\%$ below on instance 3, averaged over $K$ ($13.2\%$ at $K=1$). The lower bound supports this finding, as the decision found in $20$ minutes from one scenario is at most $0.51\%$ from the optimum on instance 1, while the solver's own solution is at most $6.43\%$ from it. For DCAP, the margin over the solver's solution ranges from $0.3\%$ to $1.6\%$, and the decisions of \guidedsurr\ are at most $1.3\%$ to $2.0\%$ from the optimum, compared to $2.0\%$ to $3.0\%$ for the solver.

This difference arises from the way the two computations scale with the number of scenarios. The extensive form includes one copy of the recourse problem for each scenario, each introducing additional integer variables that the MIP solver must handle. In contrast, the reduction methods search over the positions of the $K$ synthetic scenarios in the latent space of the autoencoder, which involves at most $192$ numbers in this benchmark, and use the $200$ empirical scenarios solely to evaluate a candidate decision through $200$ independent recourse solves in parallel. Consequently, adding scenarios increases the duration of each iteration but does not expand the search space. In problems large enough that the solver cannot close its optimality gap, this distinction determines which approach is superior, which aligns with the intended purpose of scenario reduction.

For the other four families, the exact solver either solves the extensive form or comes within approximately $1\%$ of its bound within two hours. In the EV case, which contains $7.9$ million columns but only $40$ integer variables, the solver identifies its final incumbent between two and eight minutes into the process and completes the proof of optimality after $9$ to $49$ minutes in total. The alternative methods lag behind because each iteration requires approximately $30$ seconds for recourse solves, resulting in only $43$ iterations within the computational budget. The comparison instead highlights the minimal loss incurred when the $200$ scenarios, each representing one day of hourly charging demand at $40$ locations, are reduced to a single synthetic day. Using this reduction, \guidedsurr\ concludes on average $0.59\%$ above the optimum computed with all $200$ days, and \guidedes\ is $0.54\%$ above it. All nine runs of each of the two guided estimators, \guidedsurr\ and \guidedes, end within $1\%$ of the optimum. For $K=1$, \guidedsurr\ achieves within $1\%$ after a median of $394$ seconds of combined pretraining and search, which is comparable to the time at which the exact solver attains its final incumbent.

SUC represents an intermediate case. Its extensive form contains $3.8$ million rows, and the solver needs $21$, $63$, and $7$ minutes to solve the root relaxation on instances 1, 2, and 3, respectively, before which it holds only a trivial solution of about $30$ times the optimal cost. Within the computational budget, the reduced sets are the only useful solutions available for two instances. However, once the solver identifies an incumbent, it reaches optimality, or comes within $0.003\%$ of it, within two hours, whereas the reduced sets do not. This outcome is due to two factors that make SUC the most challenging family for the search. Each iteration exceeds two minutes because every candidate decision is evaluated by solving $200$ dispatch problems for a $200$-generator system. Consequently, \guidedsurr\ completes only ten iterations and ends $13.3\%$ above the optimum at $K=1$, with its performance curve still declining in Figure~\ref{fig:exact}. A single synthetic scenario is insufficient for this family, as the gap for \guidedsurr\ decreases to $11.2\%$ at $K=2$ and $7.0\%$ at $K=3$, indicating that three synthetic scenarios halve the gap within the same $20$ minutes.

VRPSD is a problem with $8{,}460$ columns that the solver closes to within about $1\%$ of its bound in two hours. The reduced sets have limited opportunity for improvement, yet within the budget the sets of the three evolution strategies are below the solver's incumbent within the first minute on all three instances, and the solver only matches or surpasses them after $12$ to $26$ minutes. At the conclusion of two hours, the solver leads on instances 1 and 3, while the reduced sets finish $0.5\%$ and $0.03\%$ above the solver, respectively, averaging $0.18\%$ over the family at $K=1$. This represents a negligible loss when replacing $200$ scenarios with a single synthetic one.

The figures at $K=2$ and $K=3$ in Section~\ref{ec:exactK} of the e-companion provide an additional observation. When two or three synthetic scenarios are used, the first step becomes more significant because the search updates $2m$ or $3m$ coordinates simultaneously. After the initial iteration, \guidedsurr\ achieves the lowest objective among the five methods on $13$ out of $18$ instances at $K=3$ and on $12$ at $K=2$. In contrast, at $K=1$, \guidedes\ most frequently leads after the first iteration. For CFLP at $K=3$, the first reduced set of \guidedsurr\ produces a decision within $15\%$ of the reference for every instance, whereas the other four methods begin between $160\%$ and $600\%$ and need between five and twelve minutes of search to come within $15\%$ (Figure~\ref{fig:ec_exact3}). Thus, a single step along the decision sensitivity direction explores a broader latent space than eight random or heuristic probes.

Overall, a single synthetic scenario retains most of the information that $200$ scenarios provide regarding the first-stage decision. For CFLP and DCAP, the reduced sets are both faster and yield better results than the exact solver on the extensive form. In contrast, for the other four problem families, the exact solver completes, or nearly completes, within its time limit, and the reduced sets do not outperform it. However, the loss from compressing $200$ scenarios into one is minimal for EV and VRPSD, moderate for ELECTRICITY, and larger for SUC, where the loss decreases rapidly as $K$ increases. Similar to decomposition methods such as Benders decomposition \citep{benders1962partitioning}, the guided estimators, the most computationally expensive methods per iteration in Table~\ref{tab:epochs}, are justified when the extensive form is large and costly to solve. Conversely, for problems that the exact solver solves quickly, scenario reduction does not offer time savings.

The remainder of this section presents a comparison among the estimators, highlighting the differences between the methods. Section~\ref{sec:scenarios} subsequently revisits the reduced sets, which on smaller problems are the purpose of the reduction.

\subsection{What Decision-Guided Evolution Strategies Add to the Search}\label{sec:aggregate}

\subsubsection{Pooled Results}\label{sec:pooled}

Table~\ref{tab:overall} presents the pooled results for all six families, with one row per reduced-set size $K$ of $54$ runs each and a pooled row over all $162$ runs. The pooled mean gap is dominated by SUC, where the differences between methods are large, so three further measures give every run equal weight. The mean rank orders the five methods within each run from $1$ for the best to $5$ for the worst. The mean regret is the difference between a method's gap and the smallest gap among the five methods in the same run, normalized by the mean gap to the lower bound of the family so that every family has equal weight. Paired comparisons count the runs in which a method is better than, equal to, or worse than \plainm\ in the same run, and the Wilcoxon signed-rank test assesses their significance.

Across all four measures, the guided estimators outperform the baseline. Over the $162$ runs, the mean gap decreases from $4.08\%$ for \plainm\ to $1.64\%$ for \guidedes\ and $1.02\%$ for \guidedsurr. The reduction, approximately three quarters for \guidedsurr, is primarily attributable to SUC, and the median gap decreases from $0.70\%$ to $0.34\%$ and $0.41\%$. \guidedsurr\ achieves a mean rank of $2.18$ compared to $3.65$ for \plainm, and its mean regret of $0.065$ is approximately eight times lower than the $0.515$ of \plainm. The mean rank is the average rank of the Friedman test, which rejects equal performance of the five methods over the $162$ runs ($p=1.1\times10^{-23}$) and at each $K$ ($p<10^{-6}$). \guidedsurr\ outperforms \plainm\ in $112$ of the $162$ runs, matches it in $24$, and underperforms in $26$, with $p=8.7\times10^{-11}$. The ties occur when both methods select the same first-stage decision, and among the runs where the methods differ, \guidedsurr\ is superior in four out of five cases. The improvement holds at every reduced-set size, with $37$, $37$, and $38$ wins against $9$, $7$, and $10$ losses at $K=1$, $2$, and $3$, each with $p<0.01$. \guidedes\ ranks next, with a mean rank of $2.56$ and $105$ wins against $31$ losses, with $p=7.3\times10^{-9}$. Against each other, \guidedsurr\ is superior in $74$ runs, equal in $43$, and inferior in $45$, with $p=2\times10^{-4}$, as Table~\ref{tab:ec_pairs} of the e-companion shows, so the decision-focused subspace is the stronger guide when all families are taken together.

\begin{table}[t]
\TABLE
{Pooled results over the six families, one row per reduced-set size $K$ with $54$ runs each and an additional row aggregating all $162$ runs. The gap refers to the difference from the exact solver's $2$-hour incumbent and may be negative, whereas the gap to bound denotes the difference from the solver's final lower bound and is always nonnegative. Mean rank indicates the ranking of the five methods within each run, where $1$ represents the best performance, and the position orders the five methods by their mean rank at the same $K$. Mean regret is calculated as the difference between a method's gap and the best gap among the five methods in that run, normalized by the family's mean gap to the lower bound, as in Figure~\ref{fig:pooled}, and a value of $0$ indicates that the method was the best method in every run. Wins, ties, and losses are determined relative to \plainm\ within the same run, with a tie defined as achieving the same objective value within $10^{-6}$.\label{tab:overall}}
{\begin{adjustbox}{max width=\textwidth}
\begin{tabular}{lc ccc cc c ccc c}
\toprule
\multirow{2}{*}{Method} & \multirow{2}{*}{$K$} & \multicolumn{3}{c}{Gap (\%)} & \multicolumn{2}{c}{Rank} & Mean & \multicolumn{4}{c}{Paired vs.\ SPSA-DGSS} \\
\cmidrule(lr){3-5}\cmidrule(lr){6-7}\cmidrule(lr){9-12}
 & & mean & median & to bound & mean & position & regret & win & tie & loss & Wilcoxon $p$ \\
\midrule
\multirow{4}{*}{SPSA-DGSS}
 & 1 & 6.74 & 0.78 & 8.48 & 3.55 & 5 & 0.493 & --- & --- & --- & --- \\
 & 2 & 4.05 & 0.61 & 5.79 & 3.67 & 5 & 0.410 & --- & --- & --- & --- \\
 & 3 & 1.45 & 0.81 & 3.26 & 3.72 & 5 & 0.641 & --- & --- & --- & --- \\
 & all & 4.08 & 0.70 & 5.85 & 3.65 & 5 & 0.515 & --- & --- & --- & --- \\
\midrule
\multirow{4}{*}{Analytic-DF}
 & 1 & 15.86 & 1.13 & 17.60 & 3.28 & 3 & 0.843 & 20 & 8 & 26 & $0.32$ \\
 & 2 & 10.10 & 0.56 & 11.85 & 3.25 & 3 & 0.506 & 27 & 9 & 18 & $0.65$ \\
 & 3 & 7.34 & 0.66 & 9.09 & 3.43 & 4 & 0.469 & 29 & 8 & 17 & $0.42$ \\
 & all & 11.10 & 0.72 & 12.85 & 3.32 & 4 & 0.606 & 76 & 25 & 61 & $0.90$ \\
\midrule
\multirow{4}{*}{Unguided-ES}
 & 1 & 2.68 & 0.50 & 4.43 & 3.39 & 4 & 0.366 & 28 & 9 & 17 & $9.8\times10^{-3}$ \\
 & 2 & 2.00 & 0.50 & 3.75 & 3.41 & 4 & 0.395 & 26 & 11 & 17 & $0.51$ \\
 & 3 & 1.90 & 0.50 & 3.65 & 3.11 & 3 & 0.516 & 29 & 7 & 18 & $0.93$ \\
 & all & 2.19 & 0.50 & 3.94 & 3.30 & 3 & 0.426 & 83 & 27 & 52 & $0.07$ \\
\midrule
\multirow{4}{*}{Guided-DGSS-H}
 & 1 & 2.07 & 0.46 & 3.81 & 2.57 & 2 & 0.108 & 36 & 7 & 11 & $6.2\times10^{-7}$ \\
 & 2 & 1.66 & 0.31 & 3.41 & 2.66 & 2 & 0.108 & 33 & 12 & 9 & $4.3\times10^{-3}$ \\
 & 3 & 1.18 & 0.27 & 2.92 & 2.44 & 2 & 0.135 & 36 & 7 & 11 & $0.03$ \\
 & all & 1.64 & \textbf{0.34} & 3.38 & 2.56 & 2 & 0.117 & 105 & 26 & 31 & $7.3\times10^{-9}$ \\
\midrule
\multirow{4}{*}{Guided-DGSS-DF}
 & 1 & 1.59 & 0.44 & 3.33 & 2.21 & 1 & 0.053 & 37 & 8 & 9 & $6.9\times10^{-8}$ \\
 & 2 & 1.09 & 0.33 & 2.83 & 2.02 & 1 & 0.051 & 37 & 10 & 7 & $2.8\times10^{-4}$ \\
 & 3 & 0.40 & 0.42 & 2.15 & 2.30 & 1 & 0.092 & 38 & 6 & 10 & $8.9\times10^{-3}$ \\
 & all & \textbf{1.02} & 0.41 & \textbf{2.77} & \textbf{2.18} & \textbf{1} & \textbf{0.065} & \textbf{112} & 24 & 26 & $\mathbf{8.7\times10^{-11}}$ \\
\bottomrule
\end{tabular}
\end{adjustbox}}
{Wilcoxon $p$-values are two-sided. Best pooled value per column in bold.}
\end{table}

\begin{figure}[t]
\FIGURE
{\includegraphics[width=0.85\textwidth]{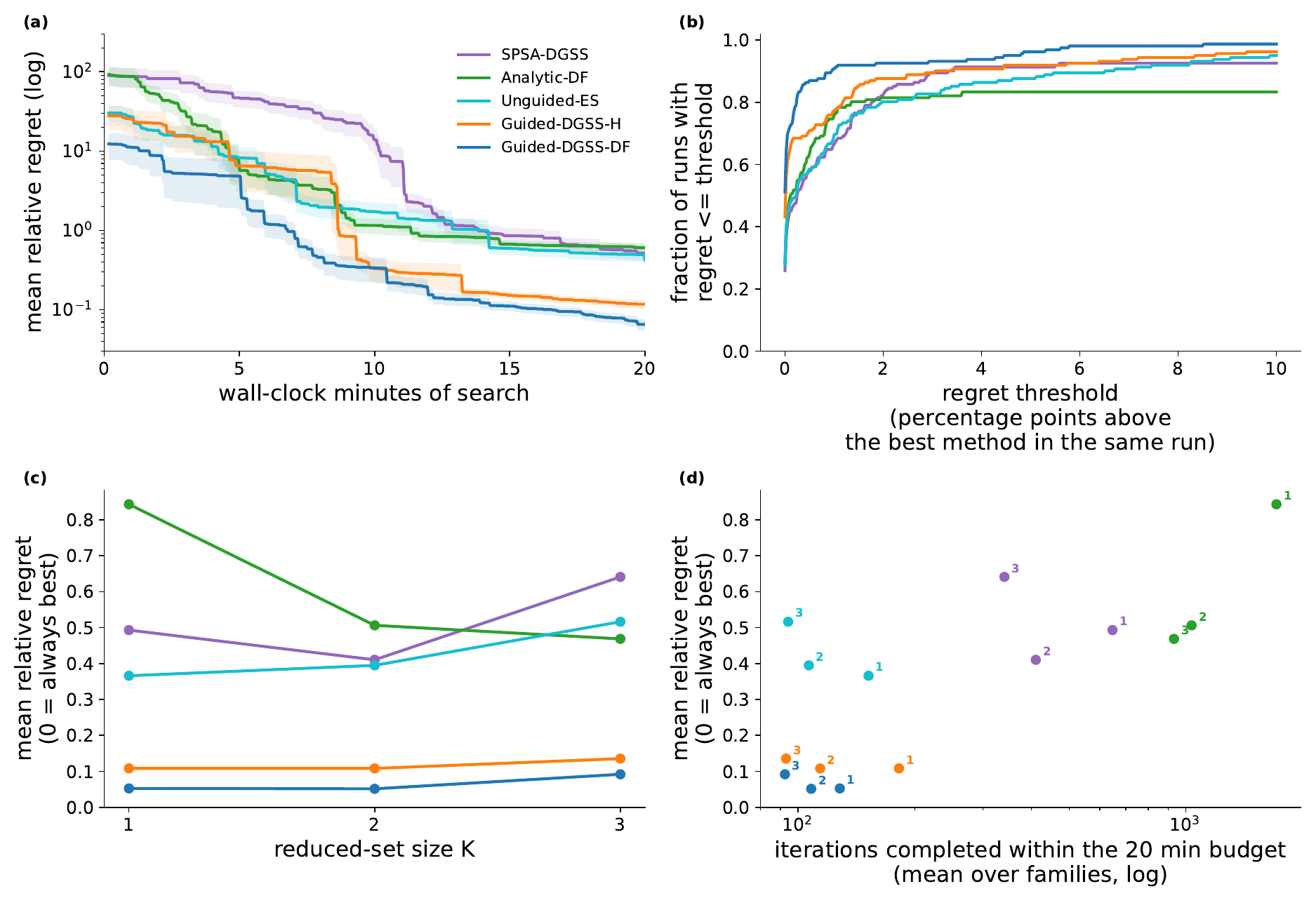}}
{Pooled results over $162$ runs per method. Regret is defined as the gap of a method in a run minus the gap of the best method in the same run, so a value of zero indicates that the method performed best. In panels (a), (c), and (d), regret is normalized by the family's mean gap to the exact solver's lower bound, ensuring comparability across families. In panel (b), regret is reported in percentage points. (a) Mean regret over the budget on a logarithmic axis, with shading representing one standard error. (b) Fraction of runs with regret at or below the threshold, where higher and earlier curves indicate better performance. (c) Mean regret as a function of reduced-set size $K$. (d) Mean regret versus iterations completed within the budget, with one point per method and $K$, where the digit denotes $K$. The lower left corner indicates better performance.\label{fig:pooled}}
{}
\end{figure}

Figure~\ref{fig:pooled} breaks the regret down over time, by reduced-set size, and against the number of iterations. Panels (a), (c), and (d) use the normalized regret of Table~\ref{tab:overall}, and panel (b) reports regret in percentage points. \guidedsurr\ has the lowest mean regret from the first minute to the end of the $20$ minutes, except for a brief interval around minute ten when \guidedes\ matches it, and it ends approximately eight times lower than \plainm\ and more than six times lower than \isoes. Stopping the search early would therefore still, on average, give the best reduced set among the five methods. In panel (b), $91\%$ of the \guidedsurr\ runs are within one percentage point of the best method, compared to $67\%$ for \plainm. Panel (c) shows that the advantage holds at every $K$. It is largest against \analytic\ at $K=1$, where a single synthetic scenario must represent the entire distribution, and against \isoes\ it grows with $K$, because the search space expands from $m$ to $3m$ coordinates and the same eight isotropic probes cover it less well. Measured against the exact solver's lower bound instead, $71\%$ of the \guidedsurr\ runs are within $2\%$ of the bound, compared to $59\%$ for \plainm\ and $57\%$ for \isoes.

\subsubsection{Cost per Iteration}\label{sec:cost}

Panel (d) of Figure~\ref{fig:pooled} and Table~\ref{tab:epochs} show what these gains cost. A multi-probe iteration evaluates $2P=16$ probes on all $200$ scenarios, whereas \plainm\ evaluates two, and with ten cores in parallel this makes an iteration between two and five times more expensive. \guidedsurr\ therefore completes between two and five times fewer iterations than the baseline, $67$ against $296$ on CFLP and $10$ against $40$ on SUC. \analytic\ is less expensive than the baseline on every family except CFLP, as its dual pass and adjoint solve require less effort than a second probe evaluation. The decision-focused guide adds the dual pass and the adjoint solve to an iteration of \isoes. The mean iteration time of \guidedsurr\ exceeds that of \isoes\ by about $20\%$ on CFLP and by about $8\%$ on EV and SUC, and it does not exceed it on the other three families, so the additional cost of the guide is small. In panel (d), the guided estimators lie in the lower left corner, so a few dozen well-directed steps outperform hundreds of single-direction steps. The trade-off is sharpest on SUC, where the better direction does not always make up for the fewer iterations, as Section~\ref{sec:perfamily} discusses.

\begin{table}[t]
\TABLE
{Iterations completed within the $20$-minute budget and mean seconds per iteration, averaged over the $27$ runs of each family. In each multi-probe iteration, $2P=16$ probes are evaluated across the $200$ scenarios, whereas a \plainm\ iteration evaluates two probes, and \analytic\ evaluates none. The mean seconds per iteration is calculated as the average over the runs of each run's iteration time. As the iteration time increases with $K$ and the number of iterations decreases, the product of these two columns exceeds $20$ minutes.\label{tab:epochs}}
{\begin{adjustbox}{max width=\textwidth}
\begin{tabular}{l ccccc ccccc}
\toprule
\multirow{2}{*}{Family} & \multicolumn{5}{c}{Iterations completed in 20 min} & \multicolumn{5}{c}{Seconds per iteration} \\
\cmidrule(lr){2-6}\cmidrule(lr){7-11}
 & SPSA & Analytic & ES & H & DF & SPSA & Analytic & ES & H & DF \\
\midrule
CFLP & 296 & 184 & 121 & 206 & 67 & 5.6 & 8.1 & 19.5 & 15.9 & 23.4 \\
DCAP & 289 & 1243 & 115 & 115 & 115 & 5.3 & 1.0 & 11.2 & 11.3 & 11.1 \\
ELECTRICITY & 1571 & 3893 & 312 & 307 & 310 & 0.8 & 0.3 & 4.0 & 4.0 & 4.0 \\
EV & 114 & 205 & 46 & 47 & 43 & 11.9 & 7.6 & 27.3 & 26.9 & 29.6 \\
SUC & 40 & 57 & 11 & 11 & 10 & 31.6 & 21.4 & 115.5 & 112.9 & 125.3 \\
VRPSD & 487 & 1781 & 100 & 93 & 112 & 4.7 & 2.1 & 16.3 & 18.1 & 14.4 \\
\bottomrule
\end{tabular}
\end{adjustbox}}
{}
\end{table}

\subsubsection{Results per Family}\label{sec:perfamily}

Table~\ref{tab:perdataset} reports the mean gap of each method per family and the paired wins, ties, and losses against \plainm, and Table~\ref{tab:ec_pairs} of the e-companion gives the comparisons between the other methods. On each of the six families, one of the two guided estimators achieves the lowest mean gap, \guidedsurr\ on four families and \guidedes\ on CFLP and EV, and \guidedsurr\ does not lose significantly to any other method on any family. On CFLP, DCAP, and VRPSD, the gap of \guidedsurr\ remains nearly constant between $K=1$ and $K=3$, so a single synthetic scenario is often sufficient to determine the decision. On ELECTRICITY, the gap decreases from $2.5\%$ to $1.6\%$, and on SUC from $13.3\%$ to $7.0\%$, so there each additional scenario reduces the remaining gap.

\begin{table}[t]
\TABLE
{Results per family, one row per reduced-set size $K$ with $9$ runs each, and a pooled row summarizing all $27$ runs. The mean gap is measured relative to the $2$-hour incumbent, and in the gap columns bold indicates the lowest value in each row. The paired columns report the number of wins, ties, and losses against \plainm\ for runs matched by instance, $K$, and seed. ES denotes \isoes, H denotes \guidedes, and DF denotes \guidedsurr. In the paired columns bold highlights the method with the highest combined number of wins and ties in the row, with ties broken by the number of wins, which does not necessarily indicate a significant difference, and underlining indicates a loss to \plainm\ that is significant at the $5\%$ level. Additional paired comparisons and $p$-values for the pooled rows are provided in Table~\ref{tab:ec_pairs} of the e-companion.\label{tab:perdataset}}
{\begin{adjustbox}{max width=\textwidth,max totalheight=0.9\textheight}
\begin{tabular}{l c ccccc cccc}
\toprule
\multirow{2}{*}{Family} & \multirow{2}{*}{$K$} & \multicolumn{5}{c}{Mean gap (\%)} & \multicolumn{4}{c}{Win / tie / loss vs.\ SPSA} \\
\cmidrule(lr){3-7}\cmidrule(lr){8-11}
 & & SPSA & Analytic & ES & H & DF & Analytic & ES & H & DF \\
\midrule
CFLP & 1 & -5.872 & -6.247 & -5.959 & \textbf{-6.338} & -6.247 & 4/5/0 & 3/5/1 & \textbf{5/4/0} & 4/5/0 \\
 & 2 & -5.423 & -5.853 & -5.925 & \textbf{-6.338} & \textbf{-6.338} & 4/3/2 & 2/5/2 & \textbf{4/5/0} & \textbf{4/5/0} \\
 & 3 & -3.244 & -5.606 & -6.014 & \textbf{-6.247} & -5.984 & 4/4/1 & 5/3/1 & \textbf{7/2/0} & 6/2/1 \\
 & all & -4.846 & -5.902 & -5.966 & \textbf{-6.308} & -6.190 & 12/12/3 & 10/13/4 & \textbf{16/11/0} & 14/12/1 \\
\addlinespace[3pt]
DCAP & 1 & -0.809 & -0.800 & -0.785 & -0.799 & \textbf{-0.815} & \textbf{3/0/6} & \underline{2/0/7} & 2/0/7 & \textbf{3/0/6} \\
 & 2 & -0.792 & -0.807 & -0.810 & \textbf{-0.821} & -0.817 & 8/0/1 & 8/0/1 & \textbf{9/0/0} & \textbf{9/0/0} \\
 & 3 & -0.793 & -0.801 & -0.819 & \textbf{-0.821} & -0.816 & 7/0/2 & \textbf{8/0/1} & \textbf{8/0/1} & \textbf{8/0/1} \\
 & all & -0.798 & -0.803 & -0.805 & -0.814 & \textbf{-0.816} & 18/0/9 & 18/0/9 & 19/0/8 & \textbf{20/0/7} \\
\addlinespace[3pt]
ELECTRICITY & 1 & 4.693 & 2.732 & 3.648 & 3.527 & \textbf{2.530} & \textbf{9/0/0} & 8/0/1 & 8/0/1 & \textbf{9/0/0} \\
 & 2 & 3.551 & \textbf{1.671} & 2.209 & 2.573 & 1.804 & \textbf{9/0/0} & \textbf{9/0/0} & \textbf{9/0/0} & \textbf{9/0/0} \\
 & 3 & 3.036 & 1.608 & 2.115 & 2.360 & \textbf{1.569} & \textbf{9/0/0} & \textbf{9/0/0} & \textbf{9/0/0} & \textbf{9/0/0} \\
 & all & 3.760 & 2.004 & 2.657 & 2.820 & \textbf{1.968} & \textbf{27/0/0} & 26/0/1 & 26/0/1 & \textbf{27/0/0} \\
\addlinespace[3pt]
EV & 1 & 0.766 & 0.991 & 1.488 & \textbf{0.537} & 0.586 & 3/0/6 & \underline{2/0/7} & 7/0/2 & \textbf{8/0/1} \\
 & 2 & 0.564 & 0.685 & 1.710 & \textbf{0.431} & 0.477 & 3/0/6 & \underline{1/0/8} & 6/0/3 & \textbf{7/0/2} \\
 & 3 & 0.712 & 0.587 & 2.258 & 0.474 & \textbf{0.467} & 6/0/3 & \underline{1/0/8} & 7/0/2 & \textbf{8/0/1} \\
 & all & 0.680 & 0.754 & 1.819 & \textbf{0.481} & 0.510 & 12/0/15 & \underline{4/0/23} & 20/0/7 & \textbf{23/0/4} \\
\addlinespace[3pt]
SUC & 1 & 41.071 & 96.834 & 17.486 & 15.303 & \textbf{13.295} & \underline{1/0/8} & 8/0/1 & \textbf{9/0/0} & 8/0/1 \\
 & 2 & 26.047 & 64.510 & 14.623 & 13.963 & \textbf{11.225} & \underline{2/0/7} & 4/0/5 & 3/0/6 & \textbf{5/0/4} \\
 & 3 & 8.419 & 47.675 & 13.702 & 11.122 & \textbf{6.983} & \underline{1/0/8} & 1/0/8 & 1/0/8 & \textbf{2/0/7} \\
 & all & 25.179 & 69.673 & 15.270 & 13.463 & \textbf{10.501} & \underline{4/0/23} & 13/0/14 & 13/0/14 & \textbf{15/0/12} \\
\addlinespace[3pt]
VRPSD & 1 & 0.569 & 1.629 & \textbf{0.178} & \textbf{0.178} & \textbf{0.178} & \underline{0/3/6} & \textbf{5/4/0} & 5/3/1 & 5/3/1 \\
 & 2 & 0.339 & 0.406 & 0.183 & 0.178 & \textbf{0.173} & 1/6/2 & 2/6/1 & \textbf{2/7/0} & 3/5/1 \\
 & 3 & 0.568 & 0.565 & 0.178 & 0.178 & \textbf{0.173} & 2/4/3 & \textbf{5/4/0} & 4/5/0 & \textbf{5/4/0} \\
 & all & 0.492 & 0.866 & 0.180 & 0.178 & \textbf{0.175} & \underline{3/13/11} & \textbf{12/14/1} & 11/15/1 & 13/12/2 \\
\bottomrule
\end{tabular}
\end{adjustbox}}
{Two-sided Wilcoxon signed-rank test on the nonzero differences.}
\end{table}

On CFLP and VRPSD, the binary first stage lets different searches converge on the same decision, so many runs tie. In the $15$ runs per family where \guidedsurr\ and \plainm\ reach different decisions, \guidedsurr\ wins $14$ and $13$, respectively. On SUC, \guidedsurr\ has a mean gap of $10.5\%$ against $25.2\%$ for \plainm, but the paired count is only $15$ wins to $12$ losses, because the outcome depends on $K$. At $K=1$, where a single synthetic scenario must be placed precisely, \guidedsurr\ wins $8$ of $9$ runs. At $K=3$, the reduced problem is less sensitive to the placement of any one scenario and the many inexpensive steps of \plainm\ win $7$ of $9$ runs, although one of its runs ends $57.6\%$ above the optimum and the mean gap of \guidedsurr\ remains lower. SUC at $K=3$ and DCAP at $K=1$ are the only settings in which \plainm\ wins the paired comparison against \guidedsurr, although not significantly, and in both the mean gap of \guidedsurr\ is still lower, $6.983\%$ against $8.419\%$ on SUC and $-0.815\%$ against $-0.809\%$ on DCAP.

\subsubsection{Source of the Improvement}\label{sec:source}

The two control methods identify where the improvement comes from. \isoes\ uses the same probes, radius, and loop as the guided estimators but no subspace. Although it evaluates eight times as many probes per iteration as \plainm, it is not significantly better, with $83$ wins and $52$ losses ($p=0.07$), and on EV it loses $23$ of $27$ runs, because its iteration costs $2.3$ times more than a \plainm\ iteration and its step does not make up for this. Both guided estimators, with the same eight probes and the same cost per iteration, outperform \isoes\ in $26$ and $27$ of the $27$ EV runs, respectively. Their subspaces are built from different information, recourse duals and evaluated costs, so on this family the concentration of the probes matters more than the specific directions. Increasing the number of probes alone therefore does not improve the search, and the improvement comes from placing the probes in the subspace.

The underlying mechanism is determined by how a step is constructed from a limited number of probes. Each antithetic probe quantifies the change in the objective along its respective direction, and the resulting step aggregates the eight probe directions, each weighted by the corresponding measured change. According to Lemma~\ref{lem:smooth} and Proposition~\ref{prop:descent}, this step, on average, points downhill on the smoothed objective for any choice of covariance. Therefore, the covariance does not influence whether the search descends on average, but rather determines how closely a single step, composed of only eight probes, approximates that average. The search operates in $d=mK$ latent coordinates, where $m$ is the latent dimension of the autoencoder specified for each family in Section~\ref{ec:arch} of the e-companion. In this benchmark, $d$ ranges from $16$ to $192$, and for CFLP and EV it is $64$ at $K=1$ and $192$ at $K=3$. \isoes\ lacks a subspace, resulting in each of its eight probes pointing in a random direction within this space, as illustrated by the isotropic search distribution in the left panel of Figure~\ref{fig:tilt}. The squared component of the gradient along a random direction is, on average, only $1/d$ of its squared length. Consequently, in $64$ coordinates, the change measured by a probe is predominantly noise, and the probe direction is largely orthogonal to the gradient. While the orthogonal components would cancel out over many probes, eight probes are insufficient, so the step of \isoes\ primarily consists of movement that does not reduce the cost. The guided estimators use the same eight probes and total variance, but the covariance \eqref{eq:sigma} allocates a proportion $1-\alpha$ of that variance to the subspace, one direction for \guidedsurr\ and four for \guidedes, thereby elongating the search distribution along the guiding directions, as depicted in the right panel of Figure~\ref{fig:tilt}. With $\alpha=0.01$, the value selected by \guidedsurr\ for all three ELECTRICITY instances, $99\%$ of the variance of each probe aligns with the decision-focused direction. When this direction closely matches the gradient, each probe measures the slope along the gradient almost directly, resulting in consistent measurements and a collective step downhill. The remaining isotropic share $\alpha$ maintains a small spread in all other directions, ensuring that a subspace oriented incorrectly slows the search but does not stop it.

\analytic\ shows what happens when the direction is followed without sampling. On ELECTRICITY, where the direction is exact, it is the second-best method of the family, closely matching \guidedsurr\ at every $K$ and outperforming all others at $K=2$. After the first five minutes, however, it improves by less than a hundredth of a percentage point on average at $K=1$, because pure descent stops at a local optimum of an objective that is not convex in the latent coordinates. \guidedsurr\ keeps improving and passes it after approximately five and a half minutes at $K=1$ and nine minutes at $K=3$. On SUC, where the direction comes from the linear relaxation, \analytic\ stalls at a gap of approximately $70\%$ above the optimum, at points that are stationary only for the relaxation, as Remark~\ref{rem:bias} anticipates. Across all families, it is not significantly better than the baseline, with a mean rank of $3.32$ compared to $3.65$ and $76$ wins against $61$ losses ($p=0.90$). Following the direction as the update is therefore unreliable where the direction is biased, whereas using it only to place the probes keeps the expected step a descent direction for any subspace by Proposition~\ref{prop:descent}. Only this combination consistently improves upon the baseline.

\subsubsection{Decision-Focused vs.\ Heuristic Guides}\label{sec:guides}

The performance of the two guides depends on the recourse structure. On the linear ELECTRICITY family, where the decision-focused direction is exact, \guidedsurr\ outperforms \guidedes\ in $25$ of $27$ runs and \isoes\ in $26$, and selects the smallest mix, $\alpha=0.01$, on all three instances. \guidedes\ selects the largest mix there, $\alpha=0.9$, and still loses to \isoes\ with $10$ wins and $17$ losses ($p=0.04$), the only family where a guided estimator is significantly outperformed by a baseline or control, because its directions tend to pull the centroids toward expensive or poorly represented scenarios rather than along the gradient. When the direction is computed from the linear relaxation of an integer reduced problem, it carries the bias described in Remark~\ref{rem:bias}, and \guidedsurr\ is never significantly worse than \guidedes. They are nearly even on EV, $14$ to $13$, and do not differ significantly on DCAP and VRPSD. On CFLP, the heuristic guide leads in six of the seven differing runs without significance, and on SUC, \guidedsurr\ outperforms \guidedes\ in $19$ of $27$ runs ($p=4\times10^{-3}$). On ELECTRICITY and SUC, the gap of \guidedsurr\ depends on the mix, and on ELECTRICITY it increases steadily with $\alpha$. On EV, placing more of the sampling along the decision-focused direction brings no further gain, as Figure~\ref{fig:ec_alpha} in Section~\ref{ec:alpha} of the e-companion shows. The EV recourse is a transportation problem whose duals can be degenerate, and its first stage is integer, so its direction is only a surrogate in the sense of Remark~\ref{rem:bias} and adds little beyond the concentration of the probes. Accordingly, the selected mix is larger where the direction is less reliable, between $0.1$ and $0.9$ on EV and between $0.3$ and $0.9$ on SUC. On CFLP, DCAP, and VRPSD, the gap of \guidedsurr\ changes little across the five values of $\alpha$, so the choice of its mix matters little on these families. The heuristic guide is therefore best regarded as a backup for the decision-focused guide when the reduced problem must be treated as a black box, as it requires neither recourse duals nor a differentiable reduced solve. Each heuristic direction is built from a single scenario, whereas the decision-focused direction aggregates the effects of all $200$ scenarios on the expected cost through their duals, which explains why the heuristic guide matches the decision-focused guide on CFLP, DCAP, EV, and VRPSD but does not significantly outperform it on any family.

\subsection{Convergence Over Time}\label{sec:budget}

\begin{figure}[p]
\FIGURE
{\includegraphics[width=0.95\textwidth]{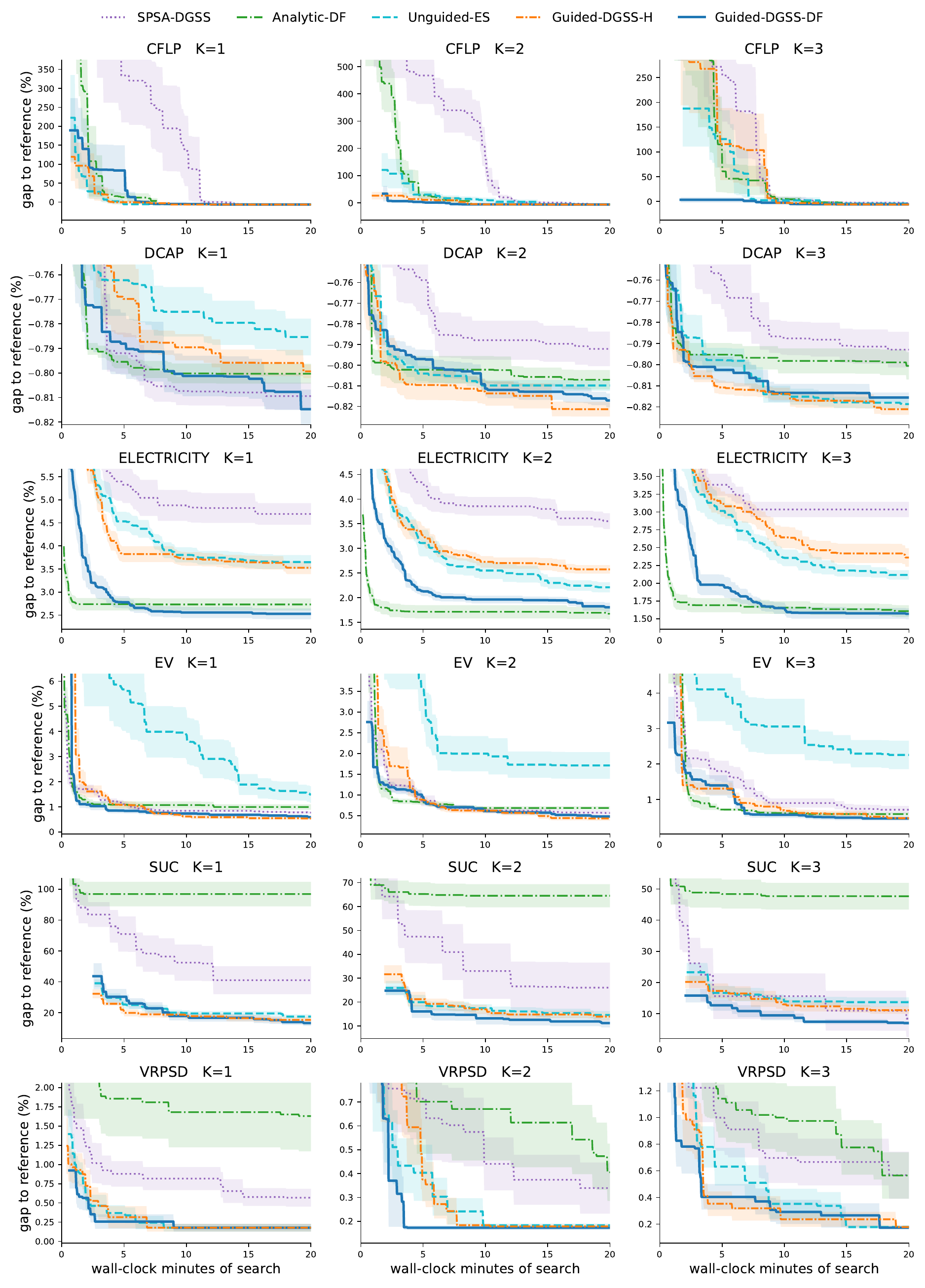}}
{Mean gap of the best solution found so far over the $20$ minutes of search, per family and $K$, over the $9$ runs of each panel. The shaded area is one standard error after centring each run on its instance mean.\label{fig:anytime}}
{}
\end{figure}

Figure~\ref{fig:anytime} presents the mean gap of the best solution found so far versus wall-clock time for each family and $K$. Each curve represents the mean over nine runs per panel, with a shaded area indicating one standard error. Three consistent patterns emerge across the panels. First, the guided estimators achieve a small gap within the initial minutes and maintain their lead. On ELECTRICITY, most runs of \plainm\ do not approach the optimum within the budget as closely as \guidedsurr\ does in its first minutes. This early advantage aligns with the variance argument in Section~\ref{sec:source}, as the initial iterations begin far from any optimal decision, making well-directed steps most impactful. Second, \analytic\ performs as predicted by theory, decreasing the gap fastest on ELECTRICITY, where its direction is exact and a single step is less costly than sixteen evaluations by other methods. On SUC, where the direction is derived from a relaxation, \analytic\ stalls within the first minutes at gaps of $48\%$ to $97\%$, and on VRPSD its gap descends slowly to $0.4\%$ to $1.6\%$, compared to under $0.2\%$ for the guided estimators. Third, in terms of mean rank, \guidedsurr\ consistently ranks first, \guidedes\ second, and \plainm\ last at $5$, $10$, $15$, and $20$ minutes, indicating that conclusions are robust to the budget cutoff. The lead of \guidedsurr\ over \plainm\ is greatest early, with $129$ of $162$ runs ahead after five minutes and $112$ at the end. All methods continue to improve in some runs during the final five minutes, \plainm\ in $35$ of $162$ runs, \isoes\ in $41$, and \guidedsurr\ in $48$, with a median improvement of $0.1$ to $0.2$ percentage points on the five families other than SUC, and $1.5$ to $3.4$ points on SUC, where the curves of all methods except \analytic\ continue to decline. With a different budget, the magnitude of differences would change. Section~\ref{ec:budget} of the e-companion demonstrates that the mean rank ordering persists when runs are extended to $30$ minutes, except that \isoes\ and \analytic\ exchange the third and fourth places, and that \plainm\ narrows its distance to \guidedsurr.

\subsection{Interpretability of the Synthetic Scenario Sets}\label{sec:scenarios}

\begin{figure}[p]
\FIGURE
{\includegraphics[width=\textwidth,height=0.90\textheight,keepaspectratio]{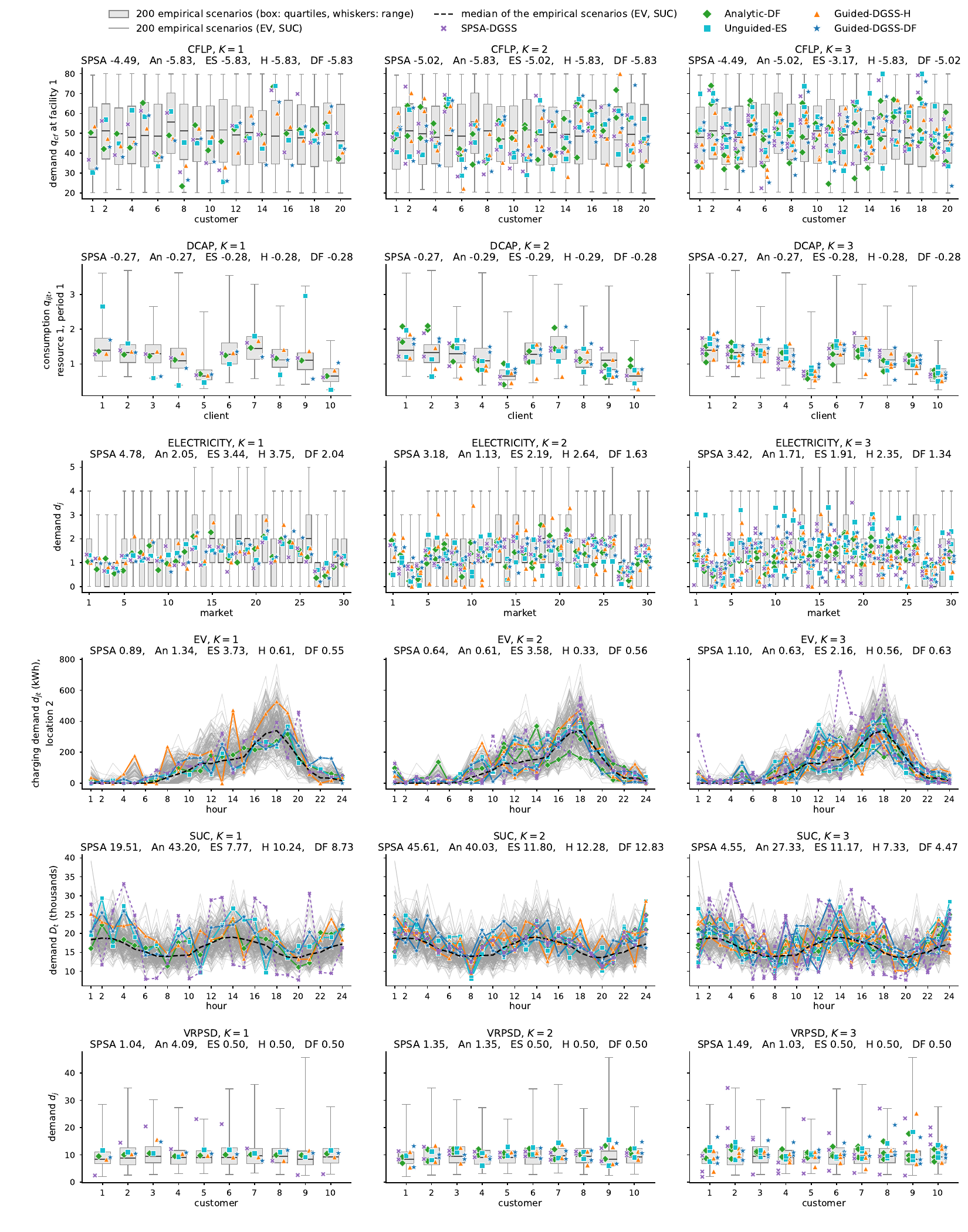}}
{Empirical scenarios (grey) and synthetic scenarios of each method in the original units, one row per family on instance 1 and one column per $K$, each drawn at a single training seed per family. Panel titles indicate the gap of each method, expressed as a percentage at $20$ minutes.\label{fig:scenmap}}
{}
\end{figure}

The gap assesses the quality of the decision rather than the scenario set that generated it. Therefore, this subsection analyzes the synthetic scenarios themselves. For one instance from each family and each $K\in\{1,2,3\}$, Figure~\ref{fig:scenmap} shows the $200$ empirical scenarios together with the synthetic scenarios to which each method converged, in the original units of the problem, with the corresponding gaps displayed above each panel. These synthetic scenarios are the decoder outputs after projection onto the data range described in Section~\ref{ec:arch} of the e-companion, representing the scenarios used to solve the reduced problem. 

A scenario has between $24$ and $960$ entries, so the figure shows the complete scenario for SUC and a part of the scenario vector for the other five families. The rows show the demand of the first $20$ customers at facility 1 for CFLP, the consumption of the ten clients at resource 1 in period 1 for DCAP, the demand of the first $30$ markets for ELECTRICITY, the hourly charging demand at location 2 for EV, the hourly demand for SUC, and the demand of the ten customers for VRPSD. For CFLP, DCAP, ELECTRICITY, and VRPSD, each box shows the quartiles of one scenario entry over the $200$ empirical scenarios, and the whiskers show its range. For EV and SUC, the grey curves are the $200$ empirical daily profiles, and the black dashed curve is their median at each hour. In the panel titles, SPSA, An, ES, H, and DF denote \plainm, \analytic, \isoes, \guidedes, and \guidedsurr.

For every problem family, the synthetic scenarios generated by the guided estimators remain within or near the range of the empirical scenarios and mostly close to the bulk of the data. On EV, for example, the synthetic days follow the empirical daily profile with its evening peak. This occurs because the training signal \eqref{eq:obj_eval} represents an expected cost over the $200$ scenarios, rather than a worst-case scenario. Consequently, the scenario identified by the search summarizes the distribution as perceived by the decision process.

Scenarios of different methods are frequently close to each other, and scenarios that are almost identical may still lead to different decisions. For example, on SUC at $K=2$, the scenarios produced by \plainm\ and \analytic\ differ by less than one percent of the data spread in each of the $24$ entries, so their curves coincide in Figure~\ref{fig:scenmap}, yet the resulting decisions differ in $130$ out of $9{,}600$ commitments, with respective gaps of $45.6\%$ and $40.0\%$. This is because the reduced decision is piecewise constant with respect to the scenarios, and a small, visually imperceptible change in a scenario can cross a discontinuity. The converse situation also arises. On VRPSD at $K=3$, the reduced sets from \isoes, \guidedes, and \guidedsurr\ all achieve the same gap of $0.50\%$, even though their scenarios differ by $7\%$ to $12\%$ of the data range on average and induce distinct first-stage solutions of equal cost. \citet{pisinger2026deep} reported similar findings within individual runs, indicating that a good reduced set is generally not unique and that the search process can arrive at an effective decision through multiple pathways.

\begin{figure}[t]
\FIGURE
{\includegraphics[width=\textwidth]{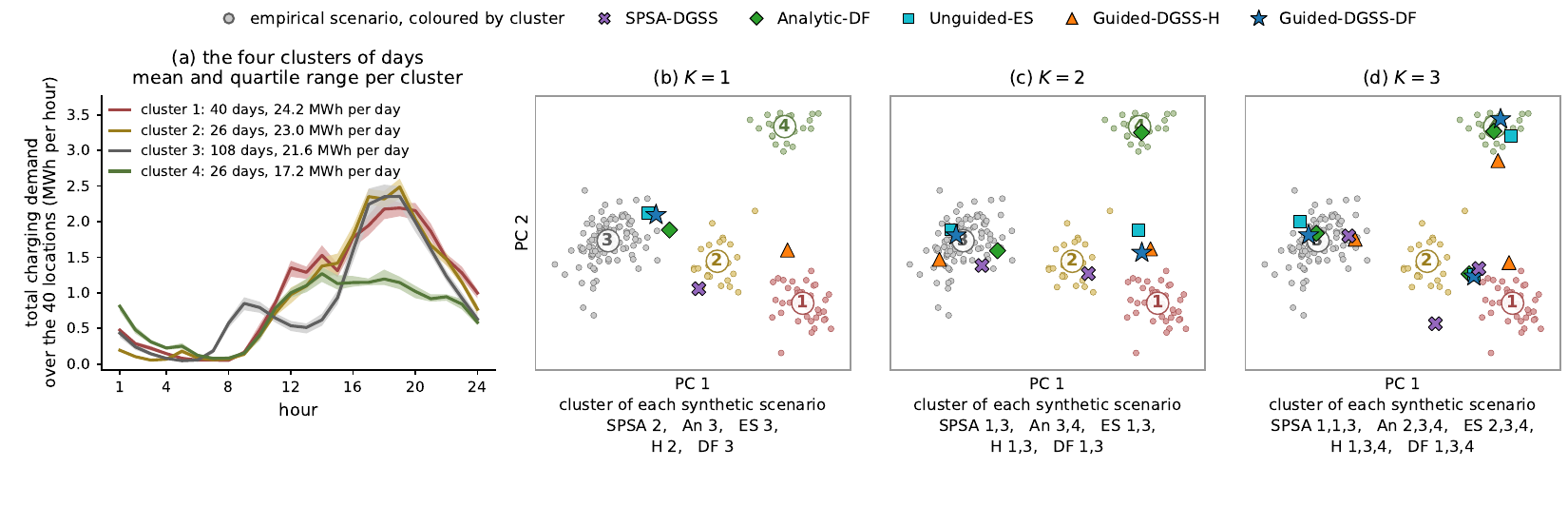}}
{Four clusters of the $200$ EV scenarios on instance 1 (a) and the synthetic scenarios of each method at $20$ minutes in the plane of the first two principal components for $K=1$, $2$, and $3$ at one training seed (b to d). Clusters are numbered from the highest to the lowest total daily demand, and the line below each panel gives the cluster of each synthetic scenario.\label{fig:evclusters}}
{}
\end{figure}

The EV family shows how the reduced set adapts to $K$. When the $200$ empirical scenarios are grouped by $k$-means in the full $960$-dimensional space, they form four clusters that differ in size and demand, corresponding to four types of days. Figure~\ref{fig:evclusters} shows these clusters, numbered from the highest to the lowest total daily demand, together with the synthetic scenarios of each method for one training seed. Cluster 3 contains $108$ of the $200$ scenarios and is therefore the most likely type of day. Across all three seeds, at $K=1$, each method places its single scenario either in cluster 3 or in cluster 2, the small cluster between the others, but never in the cluster with the highest demand or in the one with the lowest demand. At $K=2$ and at $K=3$, the reduced set contains a scenario in cluster 3 in $14$ of the $15$ runs of the five methods. At $K=3$, the three scenarios of \guidedsurr\ are distributed across three different clusters, consistently including both the largest cluster and the cluster with the lowest demand. Thus, a single scenario summarizes the typical days, and a larger reduced set captures the diversity of day types.

The search can move a synthetic scenario beyond the range of the empirical values, in which case its entries are clipped to the boundary of the range before the reduced problem is solved. In Figure~\ref{fig:scenmap}, this is the case for $79\%$ and $46\%$ of the entries of the \plainm\ scenarios on SUC at $K=1$ and $K=3$ and for $34\%$ on VRPSD at $K=3$, against at most $1\%$ for the guided estimators in the same panels. On EV, where the scenarios are only required to be positive, $8\%$ of the entries of the \plainm\ scenarios at $K=3$ lie outside the range of the empirical values. Table~\ref{tab:ec_outside} in Section~\ref{ec:outside} of the e-companion quantifies this drift for the decoder output prior to the projection onto the data range. Across all $162$ runs, \analytic\ has an average of $1.2\%$ of its entries outside the data range, \guidedsurr\ $2.5\%$, \isoes\ $3.3\%$, and \guidedes\ $3.8\%$, whereas \plainm\ exhibits $16\%$. Such drift does not inherently degrade decision quality, as the objective does not penalize scenarios outside the data range, but it affects the interpretability of the scenarios, as decision makers expect scenarios to resemble plausible days, loads, or demands. The guided estimators provide both decisions that are at least as effective as the baseline on average and reduced sets that serve as credible summaries of the data.

\section{Conclusion}\label{sec:conclusion}

Two-stage stochastic programs are often constrained more by the number of scenarios required to represent uncertainty than by the complexity of the decision itself. Scenario reduction maintains tractability, and the generative scenario search of \name{DGSS} evaluates a reduced scenario set based on the decision it induces, rather than its proximity to the original distribution, which is the relevant criterion. However, this search forfeits access to the gradient, as the mapping from synthetic scenarios to decisions involves a mixed-integer $\arg\min$. Consequently, \plainm, the estimator for \name{DGSS}, approximates the gradient by executing the entire loop at two randomly perturbed centroids and differencing the resulting costs. The first contribution of this paper is to demonstrate that the loop provides important guiding information. Each iteration inherently solves both the reduced problem and every recourse subproblem, yielding dual prices that indicate how the cost responds to the decision and how the decision responds to the scenarios. By applying the chain rule, these components can be assembled to provide the sensitivity of the full-set objective to the synthetic scenarios in closed form. Recourse duality supplies the first factor, a sparse adjoint solve of a proximally regularized relaxation provides the second, and the decoder Jacobian constitutes the third, requiring only one dual pass and one linear solve per iteration. This direction is exact when the recourse is a linear program with unique duals and the regularized relaxation is regular, with a continuous first stage ensuring that the relaxed decision is used. The approach is biased when integrality is present in either stage and unstable under degeneracy. The benchmark tests were designed to encompass all of these cases.

The second contribution addresses the question of how a search should use a direction that cannot always be trusted. Simply following the direction is intuitive but incorrect. A deterministic descent fully inherits any bias and lacks mechanisms to detect it. In the study, \analytic, which follows the direction in this manner, performs comparably to \plainm\ on the benchmark and significantly worse on SUC and VRPSD, two of the families where the direction is biased. Thus, the closed-form direction alone does not enhance the search. Its utility depends on how it is employed. The approach developed here assigns distinct roles to the two sources of information. The structure determines probe placement, while the evaluated costs of those probes decide how the search moves. Because the sampling covariance maintains an isotropic baseline, the expected update descends on the smoothed objective under any guiding subspace. Consequently, an effective guide yields a lower-variance estimate from the same evaluations, whereas an ineffective guide only reduces sampling efficiency. Bias in the structure therefore decreases the sampling efficiency and leaves the expected step a descent direction, which makes it safe to use a direction known to be inexact.

The computational study was designed to evaluate the underlying mechanism rather than solely the outcome. The unguided control employs the same estimator, probes, and radius, and it is not significantly better than the baseline, so the performance gains are attributed to probe placement rather than quantity. The deterministic control is effective when its direction is exact but can stall at a stationary point of the relaxation when it is not, as observed on SUC. In contrast, using the same direction solely to shape the sampling yields improvements over the baseline in both cases. Where the performance gap depends on the covariance mix, as observed in ELECTRICITY and SUC, the optimal mix aligns with the quality of the guide. In CFLP, DCAP, and VRPSD, the gap of the decision-focused guide is largely independent of the mix. The benefit of a well-directed step is most pronounced when a single step must traverse a significant portion of the search space, particularly at larger reduced-set sizes and during the initial minutes of each search. The reduced sets identified by the guided estimators are located within the data and near its center, as the objective is an expectation rather than a worst-case scenario. Therefore, the two or three synthetic scenarios can be interpreted as a summary of the distribution from the perspective of the decision.

The study provides four \textbf{practical guidelines}. First, apply the reduction when the extensive form is so large that the exact solver remains far from its bound within the time available, as observed in the two families with the most integer variables. In these cases, a single synthetic scenario identified in a fraction of the solver's time produced a better decision than the one found by the solver. Otherwise, use the reduction only when the reduced set itself is required, since for problems where the solver converges within minutes, the reduction does not yield time savings. Additionally, the loss in decision quality from substituting two hundred scenarios with one or three is small for most families and decreases rapidly with increased reduced-set size where it is not. Second, begin with $K=1$ and increment $K$ while the full-set cost of the induced decision, computed in each iteration, continues to decrease, as observed in SUC and ELECTRICITY but not in CFLP, DCAP, and VRPSD. Third, employ the decision-focused guide with $\alpha=0.01$ when both stages are linear and $\alpha=0.3$ otherwise, as these values are most frequently selected in Table~\ref{tab:alpha}. Use the heuristic guide only when duals are unavailable, since it is never significantly better than the decision-focused guide and underperforms relative to the unguided control when an exact gradient is available. Finally, interpret the synthetic scenarios as summaries of the distribution from the decision's perspective, rather than as worst-case scenarios.

The study leaves four directions for \textbf{future research}. First, the descent guarantee, similar to that of guided evolution strategies, applies to a single iteration in expectation, and a comprehensive convergence analysis of the entire loop, including the pullback through the autoencoder, remains outstanding. Second, the principal computational bottleneck for the guided estimators is the full-set evaluation of the probes, making a multi-probe iteration significantly more expensive than an iteration of \plainm, so reducing the computational burden of this step is the second direction. Third, the integer bias in the direction can be addressed at its origin through smoothed or perturbed optimizers. Fourth, the objective can be naturally extended from the expected cost to risk measures.

A broader implication is that using problem structure remains beneficial even when its reliability is uncertain, provided it is applied appropriately. When a direction is used for updates, the centroids move accordingly, so an inaccurate direction can lead them to suboptimal points, with no mechanism in the method to detect this error. Conversely, if the direction is used solely to determine the covariance of the probes, the update is still based on the costs returned by those probes. In this case, an inaccurate direction may result in suboptimal probe placement and increased variance in the estimate, but the expected update continues to descend. Therefore, even partially reliable structure retains value when it informs sampling rather than the update step.

\appendix
\makeatletter
\renewcommand{\@seccntformat}[1]{Appendix \csname the#1\endcsname:\ }
\makeatother
\section{Selected Covariance Mix}\label{app:alpha}

For each guided estimator and each instance, five values $\alpha\in\{0.01,0.1,0.3,0.5,0.9\}$ were tested, and the value yielding the lowest mean full-set objective across nine runs was selected. Table~\ref{tab:alpha} presents the selected values. This value is applied uniformly for every $K$ and every seed of the instance. Figure~\ref{fig:ec_alpha} in Section~\ref{ec:alpha} of the e-companion illustrates the gap at each value of $\alpha$ for each family. \guidedsurr\ selects a mix of at most $0.3$ on $14$ of the $18$ instances, including $\alpha=0.01$ on all three ELECTRICITY instances, where its direction is exact. \guidedes\ selects $\alpha\geq 0.5$ on $8$ instances, including $\alpha=0.9$ on all three ELECTRICITY instances, and therefore relies more on the isotropic component.

\begin{table}[H]
\TABLE
{The covariance mix $\alpha$ used by each guided estimator on each instance, the same for every $K$ and every seed.\label{tab:alpha}}
{\begin{adjustbox}{max width=0.9\textwidth}
\begin{tabular}{l ccc ccc}
\toprule
\multirow{2}{*}{Family} & \multicolumn{3}{c}{Guided-DGSS-DF} & \multicolumn{3}{c}{Guided-DGSS-H} \\
\cmidrule(lr){2-4}\cmidrule(lr){5-7}
 & instance 1 & instance 2 & instance 3 & instance 1 & instance 2 & instance 3 \\
\midrule
CFLP & 0.3 & 0.01 & 0.3 & 0.3 & 0.01 & 0.1 \\
DCAP & 0.3 & 0.3 & 0.3 & 0.9 & 0.3 & 0.1 \\
ELECTRICITY & 0.01 & 0.01 & 0.01 & 0.9 & 0.9 & 0.9 \\
EV & 0.1 & 0.5 & 0.9 & 0.3 & 0.01 & 0.1 \\
SUC & 0.9 & 0.3 & 0.5 & 0.5 & 0.1 & 0.9 \\
VRPSD & 0.3 & 0.01 & 0.3 & 0.5 & 0.1 & 0.5 \\
\bottomrule
\end{tabular}
\end{adjustbox}}
{}
\end{table}

\bibliographystyle{plainnat}
\bibliography{references}

\end{document}